%% file: paper_emnlp25_camera.tex
\pdfoutput=1

\documentclass[11pt]{article}

\usepackage[final]{acl}

\usepackage{times}
\usepackage{latexsym}

\usepackage[T1]{fontenc}

\usepackage[utf8]{inputenc}

\usepackage{microtype}

\usepackage{inconsolata}

\usepackage{graphicx}
\usepackage{subfigure}
\usepackage{subcaption}
\usepackage{tabularx}
\usepackage{booktabs} 
\usepackage{amsmath}
\usepackage{amssymb}
\usepackage{mathtools}
\usepackage{amsthm}

\theoremstyle{plain}
\newtheorem{example}{Example}

\theoremstyle{remark}

\newcommand{\shrink}[1]{}

\usepackage[textsize=tiny]{todonotes}

\DeclareMathOperator*{\argmax}{\arg\!\max}

\title{\textsc{FactReasoner}: A Probabilistic Approach to Long-Form Factuality Assessment for Large Language Models}

\author{
 \textbf{Radu Marinescu\textsuperscript{1}},
 \textbf{Debarun Bhattacharjya\textsuperscript{1}},
 \textbf{Junkyu Lee\textsuperscript{1}},
 \textbf{Tigran Tchrakian\textsuperscript{1}},
\\
 \textbf{Javier Carnerero Cano\textsuperscript{1}},
 \textbf{Yufang Hou\textsuperscript{1,2}},
 \textbf{Elizabeth Daly\textsuperscript{1}},
 \textbf{Alessandra Pascale\textsuperscript{1}}
\\
 \textsuperscript{1}IBM Research,
 \textsuperscript{2}IT:U - Interdisciplinary Transformation University Austria
\\
 \small{
   \textbf{Correspondence:} \href{mailto:radu.marinescu@ie.ibm.com}{radu.marinescu@ie.ibm.com}
 }
}
\begin{document}
\include{macros}

\maketitle
\begin{abstract}
    Large language models (LLMs) have achieved remarkable success in generative tasks, yet they often fall short in ensuring the factual accuracy of their outputs, thus limiting their reliability in real-world applications where correctness is critical. In this paper, we present \textsc{FactReasoner}, a novel neuro-symbolic based factuality assessment framework that employs probabilistic reasoning to evaluate the truthfulness of long-form generated responses. \textsc{FactReasoner} decomposes a response into atomic units, retrieves relevant contextual information from external knowledge sources, and models the logical relationships (e.g., entailment, contradiction) between these units and their contexts using probabilistic encodings. It then estimates the posterior probability that each atomic unit is supported by the retrieved evidence. Our experiments on both labeled and unlabeled benchmark datasets demonstrate that \textsc{FactReasoner} often outperforms state-of-the-art prompt-based methods in terms of factual precision and recall. Our open-source implementation is publicly available at: https://github.com/IBM/FactReasoner.  
\end{abstract}

\section{Introduction}
\label{sec-intro}
Large language models (LLMs) have achieved impressive improvements and demonstrated vast capabilities in recent years \cite{LLMFewShortLearner,chowdhery2022palm}, however they still struggle to guarantee the factual accuracy of the generated content. Specifically, LLMs often \emph{hallucinate}, namely they produce factual errors in which a claim contradicts well-established ground-truth knowledge \cite{zhang2023siren,sahoo2024hallu,huang2025hallu}. This makes the models unreliable in realistic situations that require factually accurate LLM-generated responses \cite{tonmoy2024hallu}.


Most modern approaches for assessing the factuality of LLM-generated long-form responses such as FactScore \cite{factscore2023emnlp}, VeriScore \cite{song2024veriscore} and others \cite{wei2024longform,bayat2025factbench} are prompt-based approaches and consist of three main stages: 1) the response is decomposed into a set of atomic units (facts or claims) which are subsequently revised or decontextualized to make them self-contained; 2) relevant evidence (or context) is retrieved for each atomic unit from an external knowledge source such as Wikipedia, and 3) each atomic unit is evaluated against the retrieved context to determine whether it is supported (factually correct) or not and a factuality score is calculated for the response. These approaches often struggle due to conflicting information between the model's internal knowledge and conflicting information within the retrieved contexts themselves \cite{factscore2023emnlp,song2024veriscore}. 

In this paper, we introduce a new perspective on long-form factuality assessment that moves beyond traditional prompt-based approaches, particularly during the evaluation phase. We propose a novel factuality assessor, \textsc{FactReasoner}, which decomposes a response into atomic units and retrieves relevant contextual evidence for each atom from an external knowledge source. Unlike prior methods that rely on prompting a language model to evaluate these atoms against the retrieved evidence, \textsc{FactReasoner} estimates the probability of each atom being supported by reasoning over a graphical model. This model encodes a joint probability distribution over the atoms and their associated contexts, constructed using probabilistic representations of entailment and contradiction relationships between the natural language utterances of the atoms and the retrieved contexts.


\textsc{FactReasoner} addresses three important limitations of the existing prompt-based approaches:

\paragraph{Context Relevance Across Atoms} In multi-atom responses, contexts retrieved for one atom can be relevant -- either supportive or contradictory -- to another atom. Prompt-based methods struggle with this, as they require saturating the model’s context window with all of the retrieved information. \textsc{FactReasoner} overcomes this limitation using a compact probabilistic representation (i.e., a graphical model) of the relationships between \emph{all} atoms in the response and \emph{all} of the retrieved contexts.

\paragraph{Handling Conflicting Contexts} Sometimes, contexts retrieved for different atoms may contradict each other.  \textsc{FactReasoner} can leverage these contradictions effectively and in a principled manner by reasoning over their probabilistic encodings which often leads to improved performance.

\paragraph{Leveraging LLM Strengths in NLI Tasks} LLMs excel at natural language inference tasks such as entailment and contradiction. \textsc{FactReasoner} builds on this strength by framing factuality assessment as a composition of these simpler tasks.

We conduct an extensive empirical evaluation on well established labeled and unlabeled benchmark datasets for long-form factuality and compare against state-of-the-art prompt-based approaches using open-source LLMs. Our results demonstrate clearly that \textsc{FactReasoner} improves significantly over its competitors in terms of factual precision and recall. We show that exploiting the relationships between the atoms and retrieved contexts, as well as between the contexts themselves, allows \textsc{FactReasoner} to identify correctly considerably more supported atoms than the competing prompt-based approaches.



The Appendix contains additional examples, experimental results and implementation details.

\section{Background} 
\label{sec-background}
We begin by providing background on graphical models and long-form factuality for LLMs. 

\subsection{Graphical Models}
A \emph{graphical model} is a tuple $\cM = \langle \bX,\bD,\bF \rangle$, where $\bX = \{X_1, \ldots, X_n\}$ is a set of variables, $\bD = \{D_1, \ldots, D_n\}$ is the set of their finite domains of values and $\bF =\{f_1, \ldots, f_m\}$ is a set of discrete positive real-valued functions. Each function $f_i$ (also called \emph{factor}) is defined on a subset of variables $\bS_i \subseteq \bX$ called its \emph{scope} and denoted by $vars(f_i)$. The function scopes of a model $\cM$ define a \emph{primal graph} whose vertices are the variables and its edges connect any two variables that appear in the scope of the same function.

The model $\cM$ defines a factorized probability distribution on $\bX$: 
\begin{align}
P(\bx) &= \frac{1}{Z}\prod_{j=1}^{m} f_j(\bx)
\end{align}

\noindent where $Z = \sum_{\bx \in \Omega(\bX)} \prod_{j=1}^{m} f_j(\bx)$ is the normalization constant $Z$ also known as the \emph{partition function} and $\Omega(\bX)$ denotes the Cartesian product of the variables domains \cite{koller2009probabilistic}.

A common inference task over graphical models is to compute the \emph{posterior marginal distributions} over all variables. Namely, for each variable $X_i \in \bX$ and domain value $x_i \in D_i$, compute:
\begin{align}\label{eq:marg}
P(x_i) &= \sum_{\bx \in \Omega(\bX)} \delta_{x_i}(\bx)\cdot P(\bx)
\end{align}

\noindent where $\delta_{x_i}(\bx)$ is $1$ if $X_i$ is assigned $x_i$ in $\bx$ and $0$ otherwise \cite{koller2009probabilistic}.

Equation \ref{eq:marg} can be solved using any probabilistic inference algorithm for graphical models, such as variable elimination \cite{dechter-book}, belief propagation \cite{pearl88,mateescu02,yedidia2005gbp}, sampling \cite{bidyuk03a,gogate07a}, search \cite{mateescu05,dechter2007aij,marinescu-aij09a}, partitioning based approximations \cite{dechter03}, or variational inference \cite{wainwright2008,liu2011}.

\subsection{Long-Form Factuality}
Let $y$ be the long-form response generated by an LLM to a query $x$. Following prior work \cite{factscore2023emnlp,song2024veriscore,wei2024longform}, we assume that $y$ can be decomposed into a set of $n$ \emph{atomic units} (or \emph{atoms}) that can be either true or false, denoted by $\cA_y = \{a_1, a_2, \ldots a_n\}$. 

An atomic unit $a_i \in \cA_y$ is defined as a short sentence conveying one piece of information. Furthermore, given an external knowledge source $\cC$\footnote{For example, $\cC$ could be Wikipedia, Google Search, or a collection of documents embedded into a vector database.}, we say that an atomic unit $a_i\in \cA_y$ is \emph{supported} by $\cC$ if there exists at least one piece of information in $\cC$ (e.g., a passage) called a \emph{context} that undebatably supports $a_i$. Otherwise, we say that the atomic unit is \emph{not supported}. 

The \emph{factual precision} $Pr(y)$ of the response $y$ with respect to a knowledge source $\cC$ is defined as: 
\begin{align}\label{eq:pr}
Pr(y) &= \frac{S(y)}{|\cA_y|}
\end{align}

\noindent where $S(y) = \sum_{i=1}^{n} \mathbb{I}[a_i \textrm{ is supported by } \cC]$ is the number of supported atomic units. 

Furthermore, the notion of \emph{factual recall} up to the $K$-th supported atomic unit denoted by $R_K(y)$ can be defined as follows: 
\begin{align}\label{eq:re}
R_K(y) = \min(\frac{S(y)}{K}, 1)
\end{align}

Combining Equations \ref{eq:pr} and \ref{eq:re} yields an $F_1$ measure for factuality denoted $F1@K$ as follows: 

\begin{align*}
    F_1@K(y) &= \left\{
    \begin {aligned}
         & \frac{2\cdot Pr(y) \cdot R_K(y)}{Pr(y) + R_K(y)} ,& S(y) > 0 \\
         & 0 ,& S(y) = 0                  
    \end{aligned}
\right.
\end{align*}

Intuitively, $F_1@K(y)$ measures the long-form factuality of a model response $y$ given the numbers of supported and not-supported atomic units in $y$. The parameter $K$ indicates the number of supported atomic units required for a response to achieve full recall \cite{wei2024longform}.


\section{The \textsc{FactReasoner} Assessor} 
\label{sec-factreasoner}

In this section, we present \textsc{FactReasoner}, a novel long-form factuality assessor that leverages probabilistic reasoning to assess the factuality of the generated response with respect to an external knowledge source $\cC$. Specifically, \textsc{FactReasoner} constructs a graphical model that captures a joint probability distribution over the atomic units in the response and their corresponding contexts in $\cC$. For each atom $a_i$, it then computes the posterior marginal probability distribution $P(a_i)$, which quantifies the likelihood that $a_i$ is true (or supported) given the information available in $\cC$.




\subsection{A Graphical Models Based Approach}
Let $y$ be the long-form response generated by an LLM for the input query $x$, and let $\cA_y = \{a_1, \ldots, a_n\}$ be the set of $n$ atomic units corresponding to $y$. For simplicity, but without loss of generality, we restrict ourselves to atomic units that are either \emph{facts} or \emph{claims} \cite{song2024veriscore}. In addition, let $\cC_y = \{c_1, \ldots, c_m\}$ be a set of $m$ contexts relevant to $y$'s atoms that were retrieved from an external knowledge source $\cC$. We make no assumptions about these contexts, namely they may be overlapping and/or contradicting each other, which is often the case in realistic scenarios. 

We next define the graphical model $\langle \bX, \bD, \bF \rangle$ that represents a joint probability distribution over the atoms and their corresponding contexts.

\paragraph{Variables.} We associate each atom $a_i \in \cA_y$ and context $c_j \in \cC_y$ with a bi-valued variable denoted by either $A_i$ (for atoms) or $C_j$ (for contexts). Therefore, we have that $\bX=\bX_a \cup \bX_c$ where $\bX_a = \{A_1, \ldots, A_n\}$ and $\bX_c = \{C_1, \ldots, C_m\}$, respectively. The domains of the variables contain the values \emph{true} and \emph{false} indicating whether the corresponding atom or context is true or false. For simplicity, we use $a_i$ and $\neg a_i$ (resp. $c_j$ and $\neg c_j$) to denote the value assignments $A_i = true$ and $A_i = false$ (resp. $C_j=true$ and $C_j=false$).

\paragraph{Priors.} For each variable $A_i \in \bX_a$ (resp. $C_j \in \bX_c$) we consider a unary factor denoted by $f(A_i)$ (resp. $f(C_j)$) representing the prior belief about the truthfulness of the corresponding atom (resp. context). Since we make no assumptions about the response, we set $f(a_i) = 0.5$ and $f(\neg a_i) = 0.5$, respectively. In contrast, the external knowledge source $\cC$ is assumed to be reliable and therefore the retrieved contexts have high probability of being true (e.g., $f(c_j) = 0.99$). Note that if a context is retrieved from a less reliable source then its prior probability can be set to a smaller value. 

\paragraph{Relationships.} In addition, we also consider binary factors denoted by $f(A_i,C_j)$ and $f(C_j, C_k)$, defined on atom-context variable pairs as well as pairs of context variables. These factors are probabilistic representations of the logical relationships between the natural language utterances corresponding to the context and atom variables. For our purpose, we use a \emph{relation model} $p_{\theta}(\cdot | t,t')$ to predict the most likely logical relationship between an ordered pair of natural language utterances from
the choices \{none, entail, contradict, equivalence\}\footnote{The ``equivalence" relationship is formed if entailment is predicted for both orderings of the utterances. The ``none" relationship corresponds to neutrality meaning that the two utterances are not related to each other.}. The relation model can be any pre-trained BERT or LLM \cite{liu2019roberta,touvron2023llama}.

Specifically, let $X$ and $Y$ be two variables in $\bX$ and let $t_X$ and $t_Y$ be their corresponding textual utterances. Let also $r^* = \argmax_{r} p_\theta(r|t_X,t_Y)$ be the predicted relationship between the ordered pair $(t_X,t_Y)$ and let $p^*$ be its probability. Table \ref{tab:factors} shows the binary factor $f(X,Y)$ corresponding to $r^*\in$ \{entailment, contradiction, equivalence\}.  

\begin{table}[t!]
\begin{center}
\resizebox{\linewidth}{!}{%
\begin{tabular}{cc|c|c|c}
        &     & entailment & contradiction & equivalence \\
    $X$ & $Y$ & $f(X,Y)$ & $f(X,Y)$ & $f(X,Y)$ \\
    \toprule
    $x$ & $y$ & $p^*$  &  $1-p^*$ &  $p^*$ \\
    $x$ & $\neg y$ & $1 - p^*$ &  $p^*$  & $1 - p^*$ \\
    $\neg x$ & $y$ & $p^*$  &  $p^*$  & $1 - p^*$ \\
    $\neg x$ & $\neg y$ & $p^*$  & $p^*$  &  $p^*$ \\
\end{tabular}}
\end{center}
\caption{Factors corresponding to logical relationships.}
\label{tab:factors}
\end{table}

For instance, if $r^*$ corresponds to entailment and $(X, Y)$ is a context-atom pair then the context supports the atom. Alternatively, if $r^*$ is a contradiction for the same $(X,Y)$ pair then the context contradicts the atom. Finally, for BERT-based relation models, the probability $p^*$ is given together with the predicted relationship $r^*$, whereas for instructed LLM-based relation models we can obtain $p^*$ by applying any uncertainty quantification (UQ) method \cite{lin2024generating,spuq}. We use a simple white-box UQ  method that calculates $p^*$ using the logits of the ``entailment" or ``contradiction" tokens produced by the model. In our experiments, we use LLM-based relation models.

Therefore, the set of factors $\bF$ is:
\begin{align*}
    \bF &= \{f(C_j,A_i)~|~ A_i\in \bX_a, C_j \in \bX_c\} \\
        & \cup \{f(C_j,C_k)~|~ C_j\in \bX_c, C_k\in \bX_c\} \\
        & \cup \{f(A_i~|~ \forall A_i\in \bX_a)\} \\
        & \cup \{f(C_j~|~ \forall C_j \in \bX_c)\}
\end{align*}
\noindent where we consider $r^*\in$ $\{\textrm{entail}$, $\textrm{contradict}\}$ for the context-atom pairs, and $r^*\in$ $\{\textrm{entail}$, $\textrm{contradict}$, $\textrm{equivalence}\}$ for the context pairs, respectively.

\begin{figure}[t!]
    \centering
    \includegraphics[width=\linewidth]{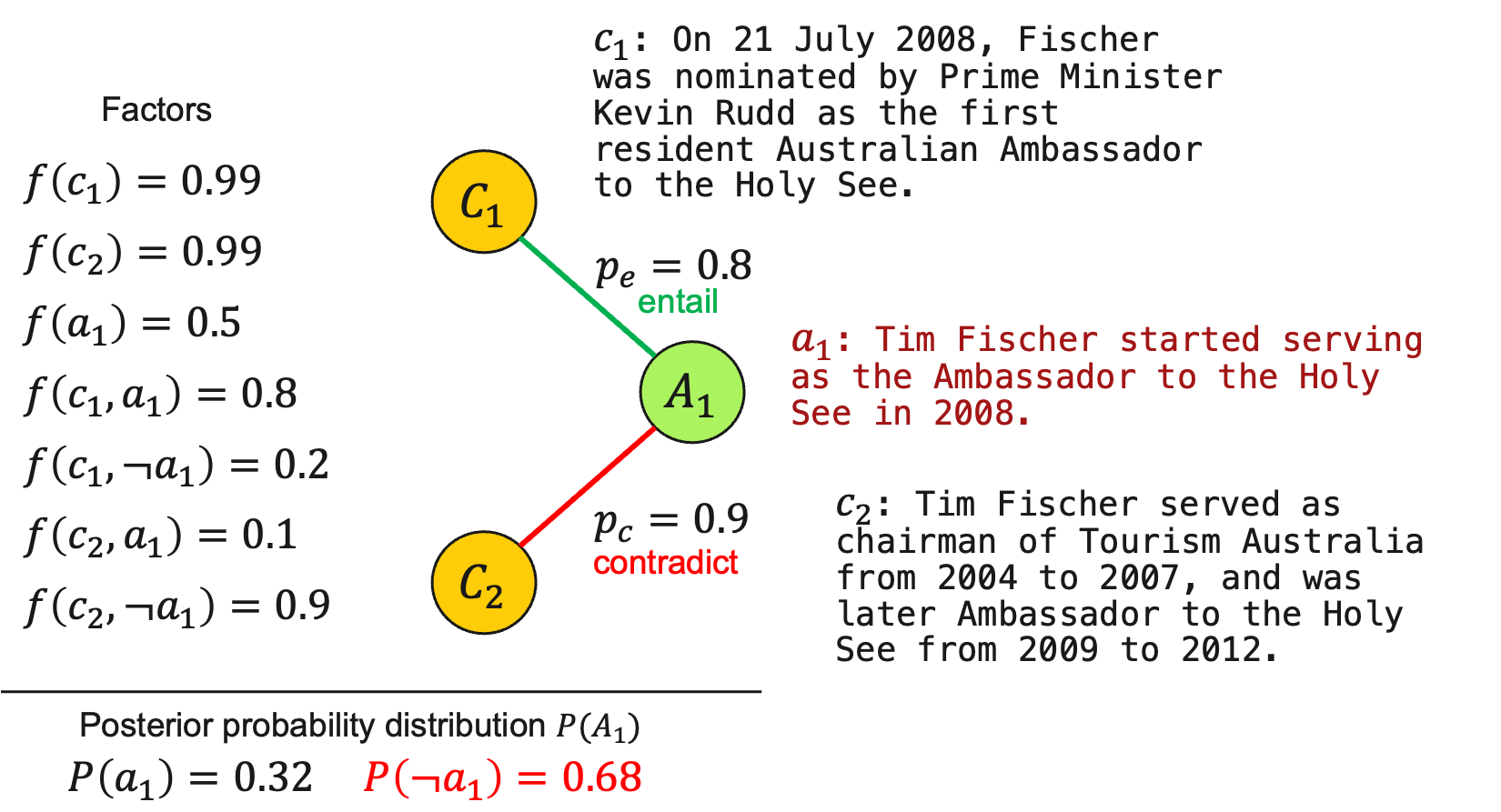}
    \caption{\textsc{FactReasoner}: the graphical model corresponding to one atom $A_1$ and two contexts $C_1$ and $C_2$ such that $C_1$ entails $A_1$ and $C_2$ contradicts $A_1$.}
    \label{fig:fr-a}
\end{figure}

\begin{example}
    Figure \ref{fig:fr-a} shows a simple example with one atomic unit $a_1$ and two contexts $c_1$ and $c_2$ retrieved from Wikipedia together with their corresponding natural language utterances. In this case, context $c_1$ entails the atom with probability $p_e=0.8$ while context $c_2$ contradicts it with probability $p_c = 0.9$. The corresponding graphical model has 3 variables $\{A_1, C_1, C_2\}$, 3 unary factors $\{f(A_1)$, $f(C_1)$, $f(C_2)\}$ as well as 2 binary factors $\{f(C_1,A_1)$, $f(C_2,A_2)\}$ encoding the two entailment and contradiction relationships.
\end{example}

\subsection{Inference and Factuality Assessment}

The graphical model $\cM =\langle \bX,\bD,\bF \rangle$ we just defined in the previous section represents a joint probability distribution over the set of atoms and relevant externally retrieved contexts. Therefore, we can use any probabilistic inference algorithm to compute the posterior marginal distribution $P(A_i)$ for each atom $A_i \in \cA_y$ \cite{pearl88,koller2009probabilistic}. Specifically, in our experiments, we use an approximate variational inference algorithm called Weighted Mini-Buckets \cite{liu2011} to compute the marginals. The algorithm is extremely efficient in practice with running times less than 0.05 seconds on all of our benchmarks.

\begin{figure}[t!]
    \centering
    \includegraphics[width=\linewidth]{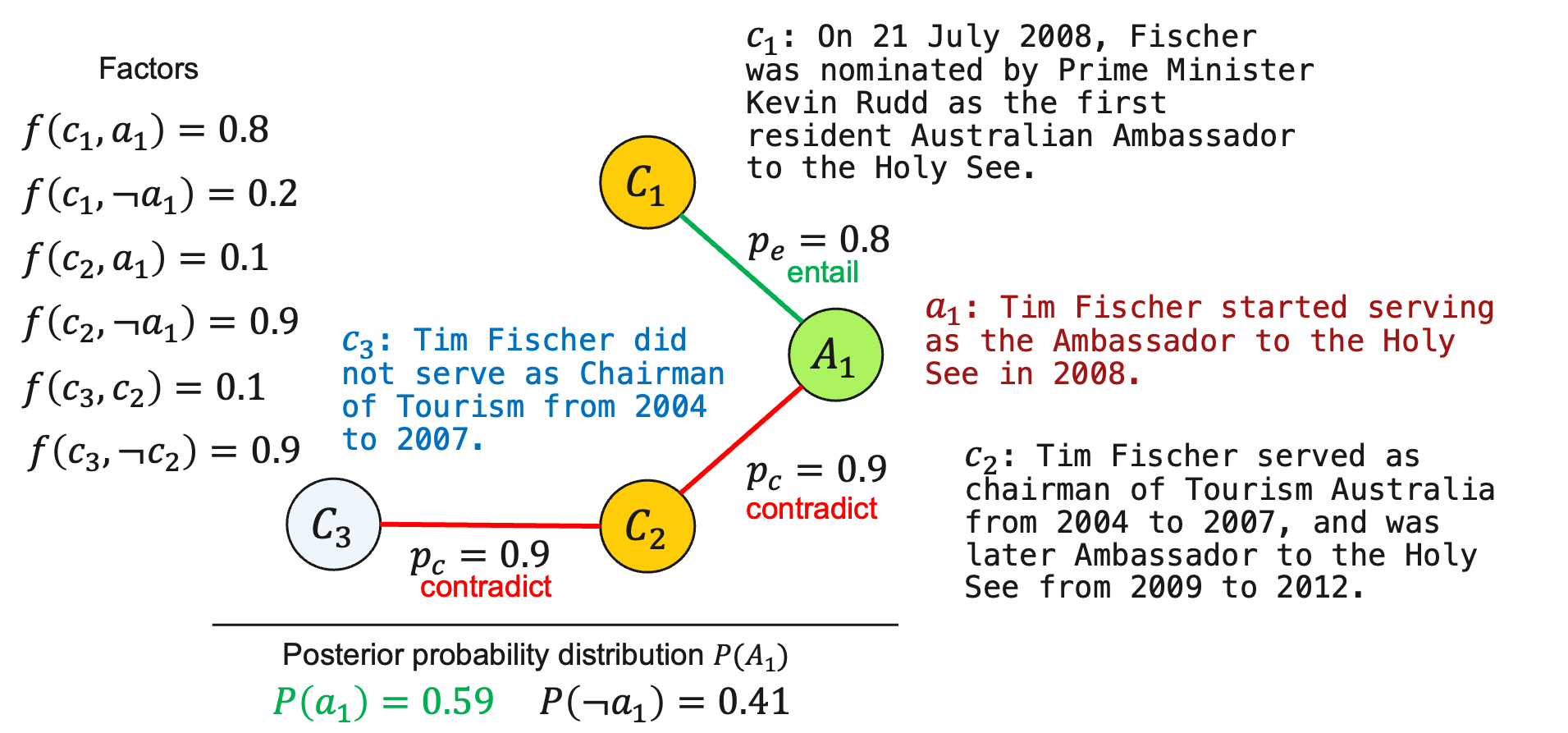}
    \caption{\textsc{FactReasoner}: the graphical model corresponding to one atom $A_1$ and three contexts $C_1$, $C_2$ and $C_3$ such that $C_3$ contradicts $C_2$.}
    \label{fig:fr-b}
\end{figure}

The number of supported atomic units $S(y)$ in a response $y$ can be computed in this case as: $S(y) = \sum_{i=1}^{n} \mathbb{I}[P(a_i) > P(\neg a_i)]$, namely it is the number of atoms for which the probability of being true is larger than the probability of being false.

\begin{example}
    Looking again at Figure \ref{fig:fr-a}, we can see that in this case the posterior probability of the atom is $P(a_1) = 0.32$ and $P(\neg a_1) = 0.68$, which means that the atom is most likely false. Figure \ref{fig:fr-b} continues the example and shows a third context $c_3$, possibly retrieved from another external knowledge source, that contradicts context $c_2$ and is neutral to atom $a_1$. As expected, the contradiction between $c_2$ and $a_1$ is much weaker now and therefore the posterior marginal probabilities are $P(a_1) = 0.59$ and $P(\neg a_1) = 0.41$, meaning that in light of the newly retrieved information, atom $a_1$ in more likely to be true than false. This example illustrates the kinds of conflicts that may exist between atoms and contexts and how they affect the factuality assessment.
\end{example}

In addition to the factual precision $Pr(y)$ and $F_1@K$ measures, we define a new entropy inspired factuality measure called $\cE(y)$ that leverages the posterior probabilities of response $y$'s atoms:
\begin{equation}\label{eq:entropy}
    \cE = \frac{1}{n} \sum_{i=1}^{n} -P(a_i) \cdot \log P(a_i)
    \vspace{-0.2em}
\end{equation}
\noindent where $n$ is the number of atomic units in $y$.

Clearly, if all atoms in $\cA_y$ have posterior probability $P(a_i) = 0.5$, there is virtually no external information to support or contradict the atoms (we refer to these atoms as \emph{undecided atoms}) then $\cE(y) = 0.150515$. On the other hand, if all atoms are true with absolute certainty ($P(a_i) = 1$), then $\cE(y) = 0$ and if all atoms are false with absolute certainty then $\cE(y) = \infty$. Therefore, when $\cE(y)$ is closer to $0$ the response is more truthful.

\subsection{The \textsc{FactReasoner} Pipeline}

\begin{figure}[t!]
    \centering
    \includegraphics[width=\linewidth]{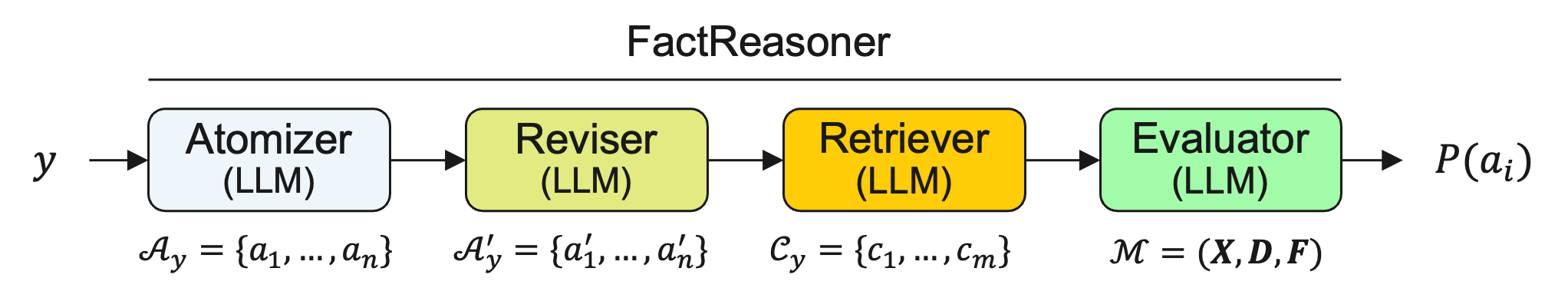}
    \caption{The \textsc{FactReasoner} pipeline.}
    \label{fig:fr-pipeline}
\end{figure}

The proposed \textsc{FactReasoner} pipeline for factuality assessment is shown in Figure \ref{fig:fr-pipeline} and consists of four main stages called Atomizer, Reviser, Retriever and Evaluator, respectively. It takes as input a response $y$ and outputs the marginal posterior probabilities $P(a_i)$ of $y$'s atomic units together with the factuality measures described earlier, such as $Pr(y)$, $F_1@K(y)$ and $\cE(y)$, respectively.

The \textbf{Atomizer} prompts an LLM to decompose the response $y$ into a set of $n$ atomic units $\cA_y$ by applying any of the decomposition strategies proposed recently \cite{factscore2023emnlp,bayat2025factbench}. Subsequently, the \textbf{Reviser} also uses an LLM to revise the atoms such that the pronouns, unknown entities, or incomplete names are replaced with their corresponding named entities in the response \cite{wei2024longform}. Next, the \textbf{Retriever} is responsible for querying an external knowledge source to retrieve the contexts relevant to the response's atoms. At this stage, we can simply use the atoms' utterances as queries or prompt an LLM to generate them \cite{song2024veriscore}. Finally, the \textbf{Evaluator} constructs the probabilistic graphical model representing the logical relationships between the atoms and contexts, and assess $y$'s factuality via probabilistic reasoning, as described previously.

Depending on what relationships between atoms and contexts are considered, we define three versions of the \textsc{FactReasoner} pipeline, as follows:

\paragraph{\textsc{FactReasoner} 1 (FR1).} In this case, for each atom variable $A_i$ up to $k$ most relevant contexts $\{C_1^i, ..., C_k^i\}$ are retrieved and only the relationships between each atom $A_i$ and its corresponding contexts are considered, namely only the factors $f(A_i,C_j^i)$ are created (where $j=1..k$).

\paragraph{\textsc{FactReasoner} 2 (FR2).} This version also retrieves up to $k$ contexts for each atom $A_i$, but it subsequently removes any duplicated contexts, thus resulting in $m$ unique contexts denoted by $\{C_1, ..., C_m\}$. It then considers the relationships between each atom $A_i$ and all $m$ contexts, creating the factors $f(A_i, C_j)$, where $j=1..m$.

\paragraph{\textsc{FactReasoner} 3 (FR3).} We consider the same contexts $\{C_1, ..., C_m\}$ as in FR2, but in addition to the atom-context relationships we also consider the context-context relationships. Thus, we create the factors $f(A_i,C_j)$ and $f(C_j,C_k)$, where $j=1..m$, $k=1..m$ and $j\neq k$, respectively.

\section{Experiments} 
\label{sec-experiments}
We empirically evaluate the \textsc{FactReasoner} assessor for long-form factuality and compare it against state-of-the-art approaches on labeled and unlabeled datasets. Although the \textsc{FactReasoner} pipeline stages can be instantiated with different LLMs, in our implementation we use the same LLM throughout the entire pipeline and focus our empirical evaluation on the \textbf{Evaluator} stage (i.e., factuality assessment). Furthermore, our open-source Python implementation is publicly available at: \href{https://github.com/IBM/FactReasoner}{https://github.com/IBM/FactReasoner}.

\paragraph{Baseline Assessors.} 
For our purpose, we consider the following state-of-the-art prompt-based long-form factuality assessors: FactScore (FS) \cite{factscore2023emnlp}, FactVerify (FV) \cite{bayat2025factbench} and VeriScore (VS) \cite{song2024veriscore}. FactScore is one of the first assessor that prompts an LLM to assess whether an atomic unit of the response is supported or not by a set of contexts relevant to the atom which are retrieved from an external knowledge source such as Wikipedia. FactVerify and VeriScore are more recent refinements of FactScore's original prompt that can accommodate other external knowledge sources such as Google Search results and enable the LLM's reasoning capabilities to evaluate the relationships between an atom and its relevant contexts. Unlike FactScore, the latter can label the atoms as supported, contradicted and undecided, respectively. In our experiments, we instantiated the competing assessors including the \textsc{FactReasoner} variants with open-source LLMs belonging to the IBM Granite\footnote{\texttt{https://huggingface.co/ibm-granite}}, Meta LLaMA\footnote{\texttt{https://huggingface.co/meta-llama}} and MistralAI Mixtral\footnote{\texttt{https://huggingface.co/mistralai}} families, namely: granite-3.0-8b-instruct, llama-3.1-70b-instruct, and mixtral-8x22b-instruct, respectively. All our LLMs are hosted remotely on compute nodes with A100 80GB GPUs and accessed via \texttt{litellm} APIs capable of serving 1500 prompts per second.

\paragraph{Datasets.} 
We experimented with the following datasets: Biographies (Bio) \cite{factscore2023emnlp}, AskHistorians (AskH) \cite{xu2023}, ELI5 \cite{xu2023}, FreshBooks (Books) \cite{song2024veriscore}, and LongFact-Objects (LFObj) \cite{wei2024longform}. These datasets have been widely adopted in prior work and are considered representative benchmarks for long-form factuality assessment, as they encompass a diverse range of topics and tasks, including creative writing, history, astronomy, chemistry, and more.

The Biographies is the only \emph{labeled} dataset available. It contains 157 biographies generated by ChatGPT for various person entities that have a Wikipedia page. Each biographic passage is also associated with a set of human generated atomic units (facts) that were labeled as \emph{supported} (S) or \emph{not-supported} (NS) by human annotators. We assume that this annotation is the ground truth.

The AskH, ELI5, Books and LFObj datasets are unlabeled and consist of collections of prompts. Specifically, the AskH and ELI5 datasets each contain 200 questions sourced from the Reddit forums r/AskHistorians and r/explainlikeimfive, respectively. The Books dataset comprises 200 paragraphs, sampled as 10 excerpts from each of 20 non-fiction books published between 2023 and 2024. Our version of the LFObj dataset is a curated subset of the original collection \cite{wei2024longform}, consisting of 10 prompts randomly selected from those related to objects spanning 38 distinct topics. For each prompt across these datasets, we generated a long-form response -- up to two paragraphs in length -- using the llama-3.3-70b-instruct model \cite{touvron2023llama}.

Additionally, we constructed a new dataset, Conflicts, comprising 1,000 claims (or atomic units) randomly sampled from the recent ConflictBank benchmark \cite{su2024conflictbank}. Each claim, originally extracted from Wikidata, is considered true (i.e., supported). For every claim, we include two associated contexts: one supporting context (the default in ConflictBank) and one conflicting context (representing misinformation, also provided by ConflictBank). Notably, these two contexts are mutually contradictory, thus offering a controlled setting for our long-form factuality evaluation.

\paragraph{Measures of Performance.} 
For each dataset $\mathcal{D}$ and each competing assessor, we report the factual precision ($Pr$) and the $F_1@K$ score, averaged over all prompts in $\mathcal{D}$. If $\mathcal{D}$ includes annotated atomic units (i.e., ground truth labels), we additionally report the standard $F_1$ score and the mean absolute error (MAE), defined as:
\begin{equation}
    \textrm{MAE} = \frac{1}{|\cD|}\sum_{j=1}^{|\cD|} |Pr_j - Pr^*_j|
\end{equation}
\noindent where $Pr_j$ and $Pr^*_j$ are the precision and the ground-truth precision for the $j$-th instance, respectively. Since the \textsc{FactReasoner} assessors calculate the posterior marginals of the atoms, we also compute the $\cE$-measure. Finally, we include the mean number of atoms classified as supported (\#S), contradicted (\#C), and undecided (\#U).


\begin{table}[t!]
    \centering
    \resizebox{0.9\linewidth}{!}{%
    \begin{tabular}{l|r|r|r|r|r}
       Dataset & \# prompts & \# atoms & \# S$^*$ & Pr$^*$ & $K$\\
       \toprule
       Biographies  & 157 & 31 & 20 & 0.62 & 32 \\
       Conflicts    & 1000 & 1 & 1 & 1.00 \\
       \midrule
       AskH         & 200 & 22 & & & 22\\
       Books        & 200 & 23 & & & 23\\
       ELI5         & 200 & 22 & & & 21\\
       LFObj        & 380 & 26 & & & 25\\
    \end{tabular}
    }
    \caption{Properties of the datasets used for evaluation.}
    \label{tab:datasets}    
\end{table}

\paragraph{External Knowledge Sources.} We consider two external knowledge sources: Wikipedia and Google Search results. For a given atom, the top $k$ results are retrieved as contexts either from wikipedia.org using the Wikipedia retriever available from LangChain\footnote{\texttt{https://python.langchain.com}}, or from google.com using the Serper API\footnote{\texttt{https://serper.dev}}. In both cases, a context is a tuple $(t,l,s,d)$, where $t$ is the title of the wiki/web-page, $l$ is the link, $s$ is a short text snippet or summary and $d$ is the content retrieved from $l$ (but capped at max 4000 characters). We used $k=3$ for the Wikipedia retriever and $k=5$ for the Google Search results \cite{factscore2023emnlp,wei2024longform}.

To ensure a consistent evaluation across all datasets, we decompose each generated response into its constituent atomic units and revise them using the same llama-3.3-70b-instruct model. Additionally, we retrieve and cache the relevant contextual information for each atom from the two designated knowledge sources. This standardized setup allows all competing assessors to be evaluated on an identical set of atoms and associated contexts. Table~\ref{tab:datasets} summarizes the key properties of the datasets, including the number of prompts, the mean number of atoms per response, and the median number of atoms ($K$), which is used in computing the $F_1@K$ metric. For the labeled datasets, we also report the true number of supported atoms ($S^*$) and ground-truth precision (Pr$^*$).


\subsection{Results on Labeled Datasets}

\begin{table}[t!]
    \centering
    \resizebox{\linewidth}{!}{%
    \begin{tabular}{l|r|r|r|r|r|r|r|r}
       Assessor & \# S & \# C & \# U & Pr $\uparrow$ & $F_1$ $\uparrow$ & $F_1@K$ $\uparrow$ & MAE $\downarrow$ & $\cE$ $\downarrow$ \\
       \toprule
       \multicolumn{9}{c}{\texttt{granite-3.0-8b-instruct}}\\
       \midrule
       FS   & 18 & 12  &     & 0.59  & 0.70  & 0.57  & 0.17 &   \\
       FV   & 14 &  2  & 14  & 0.45  & 0.67  & 0.44  & 0.21 &  \\
       VS   & 15 &  8  &  6  & 0.49  & 0.64  & 0.48  &  0.21 &  \\
       FR1 (ours)  & 14 &  2  & 14  & 0.43  & 0.70  & 0.43  & 0.22  &  0.12 \\
       FR2 (ours)  & 20 &  4  &  6  & {\bf 0.62} & {\bf 0.78}  & {\bf 0.61} & {\bf 0.12}  & {\bf 0.06} \\
       FR3 (ours)  & 19 &  4  &  6  & 0.60  & 0.78  & 0.59  & 0.13  &  {\bf 0.06} \\
       \midrule
       \multicolumn{9}{c}{\texttt{llama-3.1-70b-instruct}}\\
       \midrule
       FS   & 19 & 12  &     & 0.59  & 0.73  & 0.58  & 0.16 &  \\
       FV   & 15 & 1   & 14  & 0.47  & 0.73  & 0.47  & 0.19 &  \\
       VS   & 12 & 0   & 18  & 0.38  & 0.64  & 0.38  & 0.27 &  \\
       FR1 (ours)  & 13 & 1   & 16  & 0.42  & 0.71  & 0.42  & 0.23 & 0.10 \\
       FR2 (ours) & 19 & 2   &  9  & {\bf 0.60}  & {\bf 0.83}  & {\bf 0.59}  & {\bf 0.11} & {\bf 0.06} \\
       FR3 (ours) & 19 & 2 &  9  & {\bf 0.60}  & {\bf 0.83}  & {\bf 0.59}  & {\bf 0.11} & {\bf 0.06} \\
       \midrule
       \multicolumn{9}{c}{\texttt{mixtral-8x22b-instruct}}\\
       \midrule
       FS   & 19 & 12 &    & 0.59  & 0.74  & 0.58  & 0.16 &  \\
       FV   & 15 & 1 & 13  & 0.49  & 0.72  & 0.48  & 0.19 &  \\
       VS   & 13 & 1 & 15  & 0.42  & 0.65  & 0.42  & 0.25 &  \\
       FR1 (ours) & 14 & 0 & 15  & 0.44  & 0.72  & 0.44  & 0.21  & 0.10 \\
       FR2 (ours) & 20 & 1 &  8 & {\bf 0.63}  & {\bf 0.83}  & {\bf 0.62}  & {\bf 0.11}  & {\bf 0.07} \\
       FR3 (ours) & 20 & 1 &  9 & {\bf 0.64}  & {\bf 0.83}  & {\bf 0.62}  & {\bf 0.11}  & {\bf 0.07} \\

    \end{tabular}}
    \caption{Results on the labeled Biographies dataset using Wikipedia contexts (mean number of supported (\#S), contradicted (\#C) and undecided (\#U) atoms).}
    \label{tab:bio-wikipedia}
\end{table}

\begin{table}[t!]
    \centering
    \resizebox{\linewidth}{!}{%
    \begin{tabular}{l|r|r|r||r|r|r}
    Assessors & Pr & $F_1$ & MAE & Pr & $F_1$ & MAE \\
    \toprule
              & \multicolumn{3}{c|}{\texttt{llama-3.1-70b-instruct}} & \multicolumn{3}{c}{\texttt{mixtral-8x22b-instruct}} \\
    \midrule
       FR2 vs FS  & 0.3916 & 0.0000 & 0.0001  & 0.0421 & 0.0000 & 0.0003 \\
       FR2 vs FV  & 0.0000 & 0.0000 & 0.0000  & 0.0000 & 0.0000 & 0.0000  \\
       FR2 vs VS  & 0.0000 & 0.0000 & 0.0000  & 0.0000 & 0.0000 & 0.0000 
    \end{tabular}}
    \caption{Statistical significance tests: $p$-values for Pr, $F_1$ and MAE obtained on the labeled Biographies dataset.}
    \label{tab:bio-wiki-pvalues}
\end{table}

\paragraph{Biographies Dataset.}
Table \ref{tab:bio-wikipedia} shows the results obtained on the labeled Biographies dataset using Wikipedia retrieved contexts (the best performance is highlighted). We see that in terms or mean absolute error (MAE), precision and $F_1$ scores, the FR2 and FR3 assessors powered by stronger LLMs like llama-3.1-70b-instruct and mixtral-8x22b-instruct achieve the best performance compared to the other assessors. This is because both FR2 and FR3 can exploit the relationships between the atoms and all the retrieved contexts (as well as between the contexts themselves for FR3), not just the ones between an atom and its corresponding top $k$ contexts. Therefore, it is often the case that a context retrieved for atom $A_i$ may support or contradict another atom $A_j$ for which it wasn't retrieved. This leads to a higher number of true positives and consequently larger $F_1$ scores. We also observe that the numbers of undecided atoms is also smaller for FR2/FR3 compared with the other assessors. FR3 performs similarly to FR2 because the majority of the context-context relationships are equivalences.

When looking at the prompt-based assessors, especially FV and VS, we see that they are more conservative in terms of number of supported atoms found. This can be explained by the relatively strict instructions specified in their prompts for identifying supported/contradicted atoms. Hence the number of undecided atoms is much larger than that of FR2/FR3. The  simple prompt used by FS leads to finding a relatively large number supported atoms, across all the backend LLMs considered. However, many of these supported atoms are actually false positives which in fact is explained by the relatively smaller $F_1$ score compared with the best performing assessors FR2 and FR3, respectively. 

We observe that the lightweight FR1 assessor performs comparably to FV and VS in terms of precision, error, and $F_1$ score. This suggests that relying solely on the top-$k$ retrieved contexts to determine whether an atom is supported is inherently limited. Moreover, in cases where an atom is supported by multiple contexts but contradicted by a single, potentially spurious, context, the FR2 and FR3 assessors are able to correctly classify the atom as supported by exploiting the relative strengths of the supporting and contradicting evidence. In contrast, other assessors often misclassify such atoms as contradicted or undecided, highlighting their difficulty in resolving conflicting signals.

Additionally, we conducted one-sided $t$-tests on the $Pr$, $F_1$, and MAE metrics obtained by FR2 and its competitors, using the stronger LLaMA and Mixtral models. The resulting $p$-values are reported in Table~\ref{tab:bio-wiki-pvalues}. The near-zero $p$-values for the $F_1$ and MAE metrics indicate that FR2 significantly outperforms its competitors on these measures. In contrast, the relatively higher $p$-values observed between FR2 and FS suggest that FS exhibits higher precision but also a substantially higher false positive rate compared to FR2.

\begin{table}[t]
    \centering
    \resizebox{\linewidth}{!}{%
    \begin{tabular}{l|r|r|r}
    Assessor  & \texttt{llama (70b)} $\uparrow$ & \texttt{mixtral (22b)} $\uparrow$ & \texttt{granite (8b)} $\uparrow$ \\
    \toprule
    FS          & 0.35 & 0.74  & 0.49 \\
    FV          & 0.33 & 0.45  & \textbf{0.63} \\
    VS          & 0.06 & 0.46  & 0.56 \\
    FR1/2 (ours) & \textbf{0.88} & 0.83 & 0.61 \\
    FR3 (ours)  & 0.83 & \textbf{0.89}  & 0.62
    \end{tabular}}
    \caption{Accuracy on the labeled Conflicts dataset.}
    \label{tab:conflicts}
\end{table}

\paragraph{Conflicts Dataset.}
Table \ref{tab:conflicts} shows the accuracy -- defined as the proportion of claims correctly classified as true -- obtained by the competing assessors on the Conflicts dataset. In this case, FR1 and FR2 are identical. Notably, the FR assessors consistently outperform the prompt-based methods, particularly when leveraging more powerful models such as LLaMA or Mixtral. For instance, FR2 and FR3 -- both of which exploit the strength of supporting and conflicting relationships between the claims and their contexts -- correctly classify over 80\% of the claims when using the LLaMA model. In contrast, prompt-based approaches struggle when conflicts are present both between the contexts and between the claim and its contexts, resulting in significantly lower accuracy. These results underscore the effectiveness of the probabilistic reasoning framework employed by the proposed assessors in handling conflicts.

\begin{table}[t!]
    \centering
    \resizebox{0.9\linewidth}{!}{%
    \begin{tabular}{l|r|r|r|r|r|r}
       Assessor & \# S & \# C & \# U & Pr $\uparrow$ & $F_1@K$ $\uparrow$ & $\cE$ $\downarrow$ \\
       \toprule
       \multicolumn{7}{c}{\texttt{granite-3.0-8b-instruct}}\\
       \midrule
       FS   & 18  & 3  &   & 0.82  & 0.81  &     \\
       FV   & 14  & 1  & 7  & 0.62  & 0.62   &    \\
       VS   & 14  & 3  & 3  & 0.65  & 0.65   &    \\
       FR1 (ours)  & 13 & 4 & 4  & 0.60  & 0.60  &  0.08  \\
       FR2 (ours) & 14 & 7 & 0  & 0.63  & 0.62  &  0.04  \\
       FR3 (ours) & 15 & 7  & 0  & 0.67  & 0.66 &  0.06  \\
       \midrule
       \multicolumn{7}{c}{\texttt{llama-3.1-70b-instruct}}\\
       \midrule
       FS   & 18 & 3  &   & 0.82  & 0.80  &  \\
       FV   & 16 & 1  & 5  & 0.71  & 0.70   &  \\
       VS   & 15 & 0  & 7  & 0.66  & 0.65   &  \\
       FR1 (ours) & 12 & 1  & 8  & 0.53  & 0.54  & 0.08 \\
       FR2 (ours) & 17 & 1  & 3  & 0.76  & 0.74  & 0.04 \\
       FR3 (ours) & 17 & 2  & 3  & 0.75  & 0.74  & 0.04 \\
       \midrule
       \multicolumn{7}{c}{\texttt{mixtral-8x22b-instruct}}\\
       \midrule
       FS   & 18 & 3  &   & 0.82   & 0.80  &  \\
       FV   & 15 & 0  & 6  &  0.67  & 0.67  &  \\
       VS   & 15 & 0  & 6  &  0.68  & 0.67  &  \\
       FR1 (ours) & 14 & 0  & 8  & 0.60  & 0.60  & 0.07 \\
       FR2 (ours) & 18 & 0  & 3  & 0.80  & 0.79  & 0.04 \\
       FR3 (ours) & 18 & 0  & 3  & 0.80  & 0.79  & 0.04 \\
       \toprule
       DeepSeek-v3   &  15  &  2  &  5   &  0.69   &   0.69    & \\

    \end{tabular}}
    \caption{Results for the unlabeled AskH dataset using Google Search contexts (mean number of supported (\#S), contradicted (\#C) and undecided (\#U) atoms).}
    \label{tab:askh-google}
\end{table}

\begin{table}[t!]
    \centering
    \resizebox{0.9\linewidth}{!}{%
    \begin{tabular}{l|r|r|r|r|r|r}
       Method & \# S & \# C & \# U & Pr $\uparrow$ & $F_1@K$ $\uparrow$ & $\cE$ $\downarrow$ \\
       \toprule
       \multicolumn{7}{c}{\texttt{granite-3.0-8b-instruct}}\\
       \midrule
       FS   & 20 & 2  &    & 0.87  & 0.84   &    \\
       FV   & 16 & 0  & 6  & 0.71  & 0.70   &    \\
       VS   & 18 & 2  & 3  & 0.76  & 0.75   &    \\
       FR1 (ours)  & 18 & 0  & 3  & 0.79  & 0.77  &  0.04  \\
       FR2 (ours) & 21 & 1  & 0  & 0.90  & 0.86  &  0.02  \\
       FR3 (ours) & 17 & 5  & 0  & 0.74  & 0.72  &  0.04  \\
       \midrule
       \multicolumn{7}{c}{\texttt{llama-3.1-70b-instruct}}\\
       \midrule
       FS   & 20 & 3  &   & 0.84  & 0.82  &  \\
       FV   & 18 & 0  & 4 & 0.78  & 0.76  &  \\
       VS   & 17 & 0  & 5 & 0.72  & 0.71  &  \\
       FR1 (ours) & 14 & 1  & 7 & 0.62  & 0.62  & 0.07 \\
       FR2 (ours) & 19 & 1  & 3 & 0.80  & 0.78  & 0.04 \\
       FR3 (ours) & 18 & 2  & 2 & 0.80  & 0.78  & 0.04 \\
       \midrule
       \multicolumn{7}{c}{\texttt{mixtral-8x22b-instruct}}\\
       \midrule
       FS   & 20 & 3  &   & 0.84 & 0.82  &  \\
       FV   & 18 & 0  & 4 & 0.76 & 0.74  &  \\
       VS   & 18 & 0  & 4 & 0.79 & 0.77  &  \\
       FR1 (ours) & 16 & 0  & 6 & 0.69 & 0.68  & 0.06 \\
       FR2 (ours) & 20 & 0  & 2 & 0.86 & 0.83  & 0.03 \\
       FR3 (ours) & 20 & 0  & 2 & 0.86 & 0.83  & 0.03 \\
       \toprule
       DeepSeek-v3   &  17  &  3  &  5   &  0.72   &   0.69    & \\

    \end{tabular}}
    \caption{Results for the unlabeled Books dataset using Google Search contexts (mean number of supported (\#S), contradicted (\#C) and undecided (\#U) atoms).}
    \label{tab:books-google}
\end{table}

\begin{table}[t!]
    \centering
    \resizebox{0.9\linewidth}{!}{%
    \begin{tabular}{l|r|r|r|r|r|r}
       Method & \# S & \# C & \# U & Pr $\uparrow$ & $F_1@K$ $\uparrow$ & $\cE$ $\downarrow$\\
       \toprule
       \multicolumn{7}{c}{\texttt{granite-3.0-8b-instruct}}\\
       \midrule
       FS   & 18 & 3  &   & 0.85  & 0.84   &    \\
       FV   & 15 & 0  & 5 & 0.69  & 0.70   &    \\
       VS   & 16 & 2  & 3 & 0.71  & 0.72   &    \\
       FR1  & 14 & 3  & 3 & 0.66  & 0.67   & 0.08   \\
       FR2  & 18 & 3  & 0 & 0.82  & 0.80   & 0.03   \\
       FR3  & 16 & 3  & 0 & 0.83  & 0.82   & 0.03   \\
       \midrule
       \multicolumn{7}{c}{\texttt{llama-3.1-70b-instruct}}\\
       \midrule
       FS   & 19 & 3  &   & 0.86  & 0.84  &  \\
       FV   & 18 & 1  & 3 & 0.81  & 0.80  &  \\
       VS   & 17 & 0  & 4 & 0.78  & 0.77  &  \\
       FR1 (ours)  & 14 & 1  & 6 & 0.65  & 0.66  & 0.07 \\
       FR2 (ours) & 19 & 1  & 1 & 0.86  & 0.85  & 0.03 \\
       FR3 (ours) & 19 & 1  & 1 & 0.86  & 0.84  & 0.03 \\
       \midrule
       \multicolumn{7}{c}{\texttt{mixtral-8x22b-instruct}}\\
       \midrule
       FS   & 19 & 2  &   & 0.87   & 0.86 &  \\
       FV   & 17 & 0  & 3 & 0.79   & 0.79  &  \\
       VS   & 17 & 0  & 3 & 0.79   & 0.78  &  \\
       FR1 (ours) & 16 & 0  & 5 & 0.74   & 0.74  & 0.05 \\
       FR2 (ours) & 20 & 0  & 1 & 0.90   & 0.88  & 0.02 \\
       FR3 (ours) & 20 & 0  & 1 & 0.90   & 0.88  & 0.02 \\
       \toprule
       DeepSeek-v3   &  17  &  3  &  5   &  0.72   &   0.69    & \\

    \end{tabular}}
    \caption{Results for the unlabeled ELI5 dataset using Google Search contexts (mean number of supported (\#S), contradicted (\#C) and undecided (\#U) atoms).}
    \label{tab:eli5-google}
\end{table}

\begin{table}[t!]
    \centering
    \resizebox{0.9\linewidth}{!}{%
    \begin{tabular}{l|r|r|r|r|r|r}
       Method & \# S & \# C & \# U & Pr $\uparrow$ & $F_1@K$ $\uparrow$& $\cE$ $\downarrow$\\
       \toprule
       \multicolumn{7}{c}{\texttt{granite-3.0-8b-instruct}}\\
       \midrule
       FS   & 24 & 1  &    & 0.93  & 0.91   &    \\
       FV   & 20 & 0  & 4  & 0.79  & 0.79   &    \\
       VS   & 18 & 3  & 4  & 0.68  & 0.69   &    \\
       FR1  & 24 & 0  & 1  & 0.93  & 0.91   & 0.02   \\
       FR2  & 25 & 1  & 0  & 0.97  & 0.94   & 0.00   \\
       FR3  & 23 & 4  & 0  & 0.89  & 0.86   & 0.02   \\
       \midrule
       \multicolumn{7}{c}{\texttt{llama-3.1-70b-instruct}}\\
       \midrule
       FS   & 23 & 2  &    & 0.91  & 0.89   &  \\
       FV   & 23 & 0  & 1  & 0.91  & 0.89   &  \\
       VS   & 10 & 0  & 15 & 0.40  & 0.40   &  \\
       FR1  & 22 & 0  & 2  & 0.85  & 0.84   & 0.03 \\
       FR2  & 24 & 1  & 0  & 0.94  & 0.92   & 0.01 \\
       FR3  & 24 & 1  & 0  & 0.93  & 0.91   & 0.01 \\
       \midrule
       \multicolumn{7}{c}{\texttt{mixtral-8x22b-instruct}}\\
       \midrule
       FS   & 24 & 1  &    & 0.93   & 0.91  &  \\
       FV   & 23 & 0  & 2  & 0.90   & 0.88  &  \\
       VS   & 23 & 0  & 2  & 0.88   & 0.86  &  \\
       FR1  & 23 & 0  & 2  & 0.90   & 0.88  & 0.03 \\
       FR2  & 24 & 0  & 0  & 0.96   & 0.93  & 0.01 \\
       FR3  & 24 & 0  & 0  & 0.96   & 0.94  & 0.01 \\
       \toprule
       DeepSeek-v3   &  22  &  2  &  2   &  0.92   &   0.89    & \\

    \end{tabular}}
    \caption{Results for the unlabeled LFObj dataset using Google Search contexts (mean number of supported (\#S), contradicted (\#C) and undecided (\#U) atoms).}
    \label{tab:lfobj-google}
\end{table}

\subsection{Results on Unlabeled Datasets}
Tables \ref{tab:askh-google}, \ref{tab:books-google}, \ref{tab:eli5-google} and \ref{tab:lfobj-google} show the results obtained on the unlabeled datasets using Google Search retrieved contexts, respectively. Since there is no ground truth for these datasets, we only report the precision, $F_1@K$ (for $K=22$) and the $\cE$-measure. However, for reference, we also experimented with DeepSeek-v3 \cite{deepseek2024}, perhaps one of the strongest open models at the moment, using a suitable prompt (see Appendix).


The prompt-based assessors, FV and VS, are relatively conservative in this case and identify fewer supported atoms compared to the FR2 and FR3 assessors. In contrast, FR2 and FR3 benefit from evaluating the relationships between each atom and \emph{all} retrieved contexts, enabling them to identify more supported atoms. This advantage is reflected in their higher precision and $F_1@K$ scores. We also observe that the $\cE$-measure, specific to the FR assessors, correlates well with the number of supported atoms: as the number of supported atoms increases, $\cE$ tends to approach to $0$. Interestingly, the FS assessor, identifies more supported atoms than any other method. However, we hypothesize that a portion of these atoms may be false positive -- an issue observed in the labeled datasets. Nonetheless, in the absence of ground-truth annotations, this hypothesis remains difficult to verify.

Compared to DeepSeek-v3, FV and VS yield very similar results -- likely due to the similarity of their prompts. In contrast, FR2/FR3 identify slightly more supported atoms, though the difference is minimal. This is because some contexts support atoms they weren’t originally retrieved for.  

In summary, our proposed \textsc{FactReasoner} assessor achieved the best performance on the labeled datasets, nearly matching the ground truth. However, on the unlabeled datasets, its performance was comparable with that of its competitors including DeepSeek-v3, a very powerful open model. 

\section{Related Work} 
\label{sec-related}

Factuality evaluation of LLMs has received growing attention due to their widespread use. Early benchmarks such as TruthfulQA \cite{lin2022truthfulqa}, FreshQA \cite{vu2023freshllms}, HaluEval \cite{li2023halueval}, HalluQA \cite{cheng2023evaluating}, and FELM \cite{chen2023felm} focus on short-form factuality, assessing isolated factoids. More recent work \cite{factscore2023emnlp,wei2024longform,bayat2025factbench,song2024veriscore} extends this to long-form responses by decomposing them into atomic facts evaluated against external evidence -- typically assuming non-conflicting sources.

However, conflicting information is common in real-world knowledge bases \cite{xu2024knowledge}, posing challenges for retrieval-augmented generation systems \cite{lewis2021retrievalaugmented}. New benchmarks have emerged to capture such conflicts more realistically \cite{hou2024wiki,marjanovic2024dynamicqa,su2024conflictbank,pham2024s}.

Our work is also related to recent efforts on improving self-consistency in LLMs through formal reasoning \cite{wang2023selfconsistency,dohan2022language,mitchell2022concord}.

\section{Conclusion}
\label{sec-conclusion}

This paper introduces a new approach to long-form factuality assessment through \textsc{FactReasoner}, a novel assessor that leverages probabilistic reasoning to evaluate the factual accuracy of LLM-generated responses. Like existing prompt-based methods, \textsc{FactReasoner} decomposes responses into atomic units and retrieves relevant contexts from an external knowledge source. However, it goes further by modeling the logical relationships between atoms and contexts using a graphical model, enabling more robust factuality judgments. Experiments on both labeled and unlabeled benchmarks show that \textsc{FactReasoner} significantly outperforms existing prompt-based approaches. 


\section*{Limitations}
We acknowledge further limitations of the proposed FactReasoner approach. 

First, the Atomizer stage is sensitive to the quality of the prompt and few shot examples used as well as the LLM employed to perform the atomic unit decomposition of the response. In our work we only consider open-source models from the LLaMA family (i.e., \texttt{llama-3.3-70b-instruct}). Furthermore, the decomposition of the response can be done at different granularities such as sentence level, paragraph level and the entire response level. Our implementation is limited to decomposing the entire response in one shot.

Second, the Reviser stage is also sensitive to how well the prompt is crafted as well as the quality of the few shot examples included in the prompt. Again, at this stage we only used the \texttt{llama-3.3-70b-instruct} model.

Third, the quality of the contexts retrieved for each atomic unit depends on the implementation of the retriever used as well as the structure of the query string that it receives. Our implementation is limited to off-the-shelf retrievers such as the one available from LangChain and we used the atomic unit's utterance as query. It is possible to prompt an LLM to generate better quality queries as suggested in previous work \cite{song2024veriscore}. Therefore, employing a more advanced retriever will lead to better quality retrieved contexts and consequently will improve the overall performance of the proposed FactReasoner assessors.

Fourth, extracting the logical relationships between atoms and contexts as well as between the contexts themselves also depends on the quality of the prompt and the LLM. As before, for our relation model we only used open-source models such as granite-3.0-8b-instruct, llama-3.1-70b-instruct, and mixtral-8x22b-instruct with a fairly straightforward prompt. It is possible to craft better prompts that could lead to a better extraction of the relationships. Fine-tuning is another option to obtain a stronger relation model.

Finally, from a computational overhead perspective, the FR3 version requires $O(n\cdot m + m^2)$ prompts to extract the relationships between atoms and context, the FR2 version requires $O(n\cdot m)$ prompts while FR1 requires $O(k\cdot n)$ prompts, where $n$ is the number of atomic units, $m$ is the total number of non-duplicated contexts retrieved for the atoms, and $k$ is maximum number of contexts retrieved per atom. In contrast, the prompt-based factuality assessor only require $O(n)$ prompts.

\section*{Ethical Statement}
We recognize the positive and negative societal impacts of LLMs in general, including potential misuse of our work around uncertainty quantification for LLM generated output. We note that the datasets considered are public and peer reviewed, there are no human subjects involved, and as far as we know, there are no obvious harmful consequences from our work. All creators and original owners of assets have been properly credited and licenses and terms of use have been respected. We have not conducted crowd-sourcing experiments or research with human subjects.

\bibliography{ref}

\appendix

\section{Details on Graphical Models}
\label{apx:pgm}
Graphical models such as Bayesian or Markov networks provide a powerful framework for reasoning about conditional dependency structures over many variables \cite{pearl88,koller2009probabilistic}. 

A \emph{graphical model} is a tuple $\cM = \langle \bX,\bD,\bF \rangle$, where $\bX = \{X_1, \ldots, X_n\}$ is a set of variables, $\bD = \{D_1, \ldots, D_n\}$ is the set of their finite domains of values and $\bF =\{f_1, \ldots, f_m\}$ is a set of discrete positive real-valued functions. Each function $f_i$ (also called \emph{factor}) is defined on a subset of variables $\bS_i \subseteq \bX$ called its \emph{scope} and denoted by $vars(f_i)$. The model $\cM$ defines a factorized probability distribution on $\bX$: 
\begin{align}\label{eq:pgm}
P(\bx) = \frac{1}{Z}\prod_{j=1}^{m} f_j(\bx) ~\textrm{s.t.}~ & Z = \sum_{\bx \in \Omega(\bX)} \prod_{j=1}^{m} f_j(\bx)
\end{align}
\noindent where the normalization constant $Z$ is known as the \emph{partition function} and $\Omega(\bX)$ denotes the Cartesian product of the variables domains.

The function scopes of a model $\cM$ define a \emph{primal graph} whose vertices are the variables and its edges connect any two variables that appear in the scope of the same function.

A common inference task over graphical models is to compute the posterior marginal distributions over all variables. Namely, for each variable $X_i \in \bX$ and domain value $x_i \in D_i$, compute:
\begin{align}\label{eq:marginal}
P(x_i) &= \sum_{\bx \in \Omega(\bX)} \delta_{x_i}(\bx)\cdot P(\bx)
\end{align}
\noindent where $\delta_{x_i}(\bx)$ is $1$ if $X_i$ is assigned $x_i$ in $\bx$ and $0$ otherwise \cite{koller2009probabilistic}.

\begin{figure}
    \centering
    \includegraphics[width=1.0\linewidth]{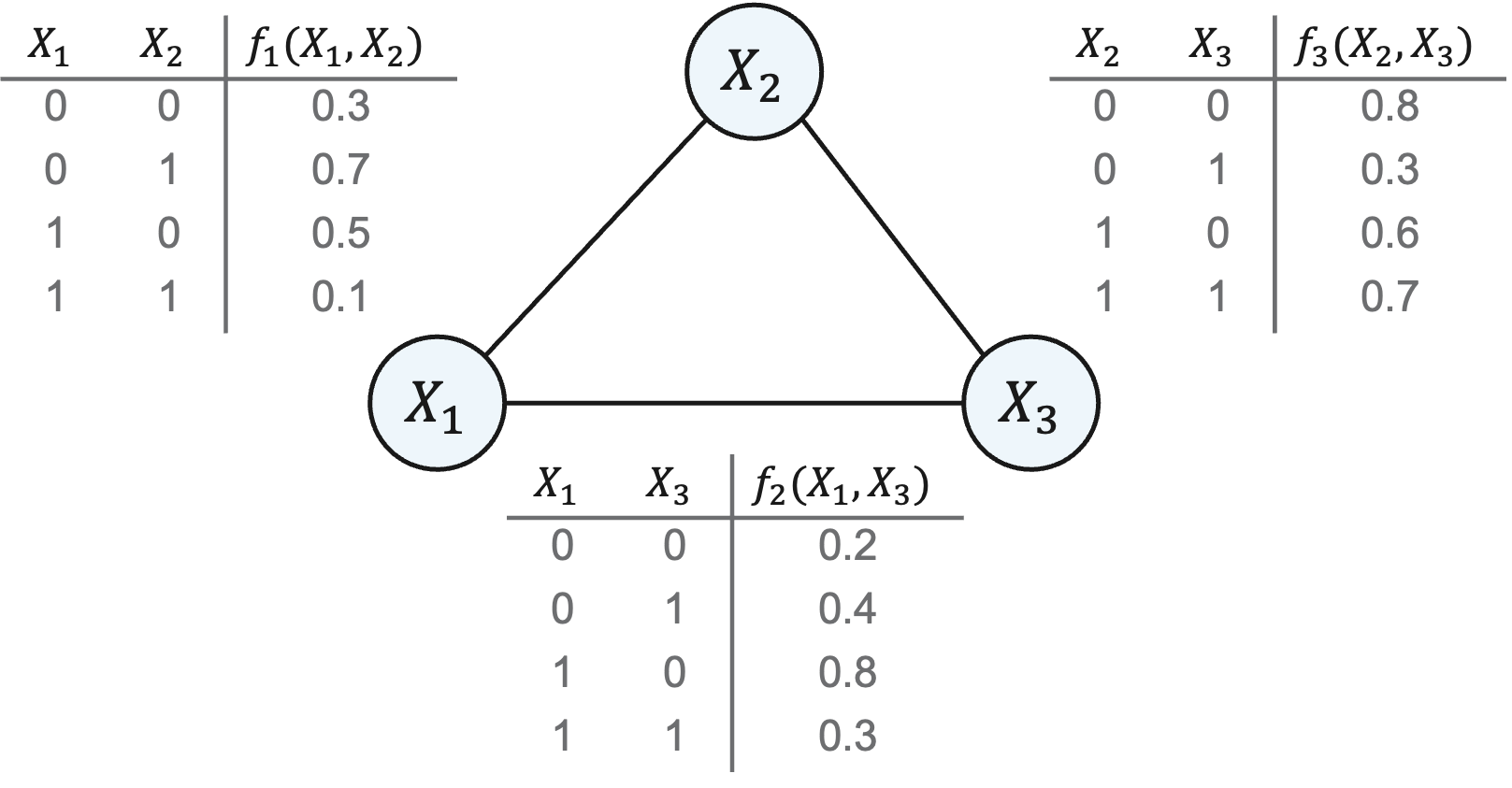}
    \caption{A graphical model with three bi-valued variables $X_1$, $X_2$ and $X_3$, and three binary functions.}
    \label{fig:pgm}
\end{figure}

\begin{example}
 Figure \ref{fig:pgm} shows a graphical model with 3 bi-valued variables $X_1$, $X_2$ and $X_3$ and 3 binary functions $f_1(X_1,X_2)$, $f_2(X_1,X_3)$ and $f_3(X_2,X_3)$. The joint probability distribution is given by $P(X_1,X_2,X_3)=\frac{1}{Z} \cdot f_1(X_1,X_2)\cdot f_2(X_1,X_3)\cdot f_3(X_2,X_3)$. In this case, the posterior marginal distribution of $X_1$ is: $P(X_1=0) = 0.46$ and $P(X_1=1) = 0.54$, respectively. 
\end{example}

Equation \ref{eq:marginal} can be solved using any probabilistic inference algorithm for graphical models, such as variable elimination \cite{dechter-book}, belief propagation \cite{pearl88}, or variational inference \cite{liu2011}. In our implementation, we employed the Weighted Mini-Buckets (WMB) algorithm \cite{liu2011}. WMB is parameterized by an i-bound, which controls the trade-off between computational complexity and inference accuracy. For our experiments, we selected an i-bound of 6, which enabled us to solve all inference problems efficiently. Notably, WMB proved highly effective in practice, solving each inference instance in under 0.05 seconds across our benchmark datasets.

\section{Details on Long-Form Factuality Assessment}
Assessing the factuality of long form text generations is a challenging problem because these kinds of generations may contain a large number of informative statements and validating each piece of information against one or more reliable sources may be time-consuming, costly and often prone to errors \cite{factscore2023emnlp,wei2024longform}.

Formally, let $y$ be the long form text generated by a large language model $\cL$ in response to a query $x$. Following prior work \cite{factscore2023emnlp,song2024veriscore}, we assume that $y$ consists of $n$ \emph{atomic units} (or \emph{atoms}) that can be either true or false, denoted by $\cA_y = \{a_1, a_2, \ldots a_n\}$. An atomic unit $a_i \in \cA_y$ is defined as a short sentence conveying one piece of information. Furthermore, given an external knowledge source $\cC$\footnote{For example, $\cC$ could be Wikipedia, the Web, or a collection of documents embedded into a vector database.}, we say that an atomic unit $a_i\in \cA_y$ is \emph{supported} by $\cC$ if there exists at least one piece of information in $\cC$ (e.g., a passage) called a \emph{context} that undebatably supports $a_i$. Otherwise, we say that the atomic unit is \emph{not supported} \cite{factscore2023emnlp,song2024veriscore}.

Therefore, the \emph{factual precision} $Pr(y)$ of the response $y$ with respect to a knowledge source $\cC$ is defined as:
\begin{align}\label{eq:precision}
Pr(y) &= \frac{S(y)}{|\cA_y|}
\end{align}
\noindent where $S(y) = \sum_{i=1}^{n} \mathbb{I}[a_i \textrm{ is supported by } \cC]$ is the number of supported atomic units.
Similarly, the notion of \emph{factual recall\footnote{Measuring recall is quite challenging because it is almost impossible to come up with a definite set of atomic units that should be included in a long form response \cite{wei2024longform}} up to the $K$-th supported atomic unit} denoted by $R_K(y)$ can be defined as follows:
\begin{align}\label{eq:recall}
R_K(y) &= \min(\frac{S(y)}{K}, 1)
\end{align}

Combining Equations \ref{eq:precision} and \ref{eq:recall} yields an $F_1$ measure for factuality denoted $F1@K$ as follows:
\begin{align}\label{eq:f1k}
    F_1@K(y) &= \left\{
    \begin {aligned}
         & \frac{2\cdot Pr(y) \cdot R_K(y)}{Pr(y) + R_K(y)} ,& S(y) > 0 \\
         & 0 ,& S(y) = 0                  
    \end{aligned}
\right.
\end{align}

Intuitively, $F_1@K(y)$ measures the long-form factuality of a model response $y$ given the numbers of supported and not-supported atomic units in $y$. The parameter $K$ indicates the number of supported atomic units required for a response to achieve full recall \cite{wei2024longform}.

The precision and recall definitions however assume that the pieces of information in $\cC$ do not conflict or overlap with each other \cite{factscore2023emnlp}.


\begin{figure}
    \centering
    \includegraphics[width=1.0\linewidth]{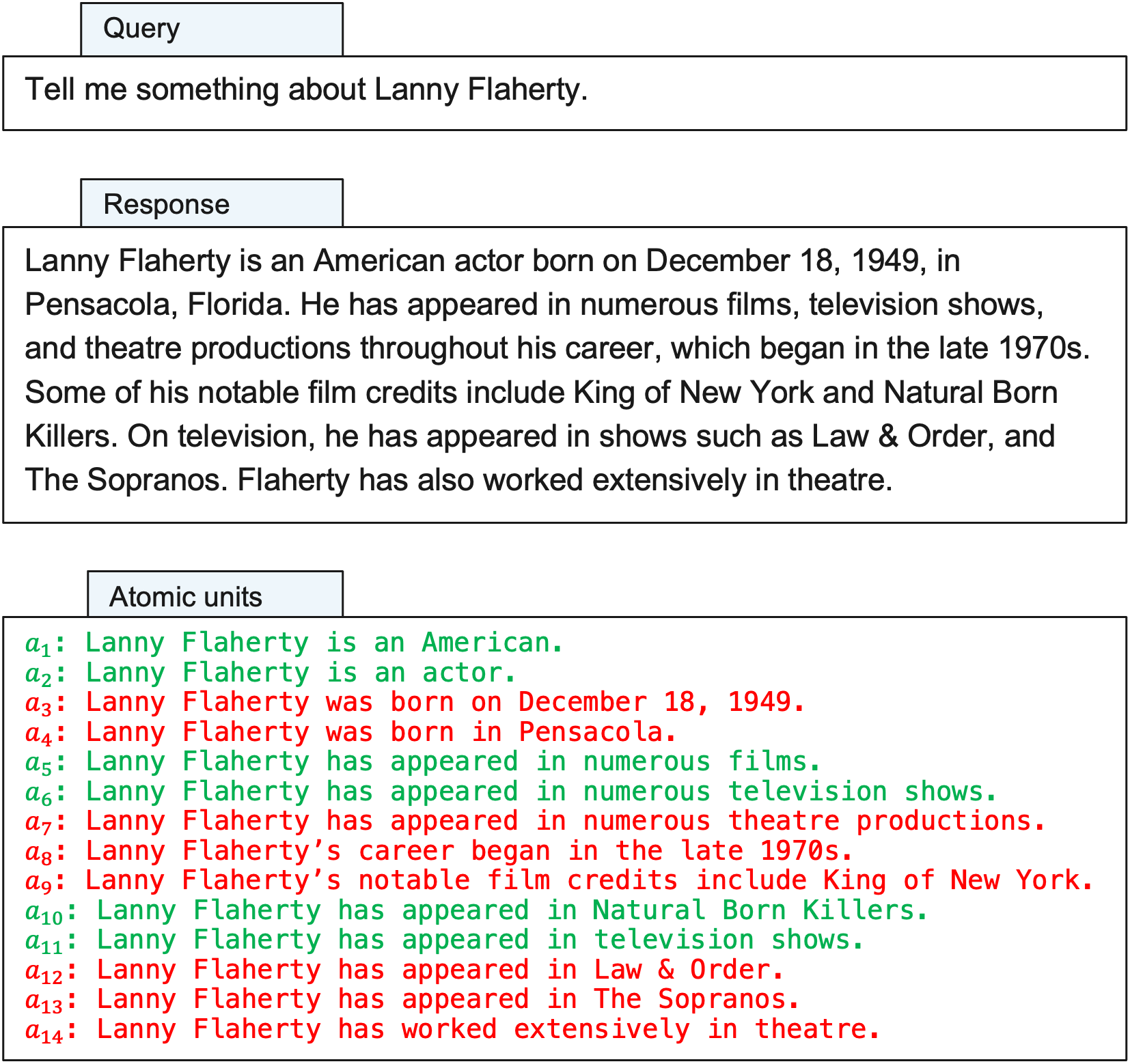}
    \caption{An example user prompt and the corresponding long form response together with its supported (green) and not supported (red) atomic units.}
    \label{fig:flaherty}
\end{figure}

\begin{example}
    In Figure \ref{fig:flaherty} we show an example of a long form generated text for a user prompt/query. In this case, the response $y$ contains 14 atomic units $\cA_y = \{a_1, a_2, \ldots, a_{14}\}$. Furthermore, considering Wikipedia as our reliable knowledge source, we depict in green the supported atomic units, while the ones in red are not supported. The factual precision and $F_1@K$ of the response are $Pr(y)= 0.43$ and $F_1@K(y) = 0.57$ for $K=7$, respectively.
\end{example}

\section{Additional Experiments}
\label{apx:experiments}

In this section, we empirically evaluate our proposed \textsc{FactReasoner} assessor for long-form factuality and compare it against state-of-the-art approaches on labeled and unlabeled datasets. Although the \textsc{FactReasoner} pipeline stages can be instantiated with different LLMs, in our implementation we use the same LLM throughout the entire pipeline and focus our empirical evaluation on the \textbf{Evaluator} stage (i.e., factuality assessment).

\paragraph{Baseline Assessors.} 
For our purpose, we consider the following state-of-the-art prompt-based long-form factuality assessors: FactScore (FS) \cite{factscore2023emnlp}, FactVerify (FV) \cite{bayat2025factbench} and VeriScore (VS) \cite{song2024veriscore}. FactScore is one of the first assessor that prompts an LLM to assess whether an atomic unit of the response is supported or not by a set of contexts relevant to the atom which are retrieved from an external knowledge source such as Wikipedia. FactVerify and VeriScore are more recent refinements of FactScore's original prompt that can accommodate other external knowledge sources such as Google Search results and enable the LLM's reasoning capabilities to evaluate the relationships between an atom and its relevant contexts. Unlike FactScore, the latter can label the atoms as supported, contradicted and undecided, respectively. In our experiments, we instantiated the competing assessors including the FactReasoner variants with open-source LLMs belonging to the IBM Granite\footnote{\texttt{https://huggingface.co/ibm-granite}}, Meta Llama\footnote{\texttt{https://huggingface.co/meta-llama}} and MistralAI Mixtral\footnote{\texttt{https://huggingface.co/mistralai}} families, namely: granite-3.0-8b-instruct, llama-3.1-70b-instruct, and mixtral-8x22b-instruct, respectively. All our LLMs are hosted remotely on compute nodes with A100 80GB GPUs and accessed via \texttt{litellm} APIs.

\paragraph{Datasets.} 
We experimented with the following datasets: Biographies (Bio) \cite{factscore2023emnlp}, AskHistorians (AskH) \cite{xu2023}, ELI5 \cite{xu2023}, FreshBooks (Books) \cite{song2024veriscore}, and LongFact-Objects (LFObj) \cite{wei2024longform}. These datasets have been widely adopted in prior work and are considered representative benchmarks for long-form factuality assessment, as they encompass a diverse range of topics and tasks, including creative writing, history, astronomy, chemistry, and more.

The Biographies is the only \emph{labeled} dataset available. It contains 157 biographies generated by ChatGPT for various person entities that have a Wikipedia page. Each biographic passage is also associated with a set of human generated atomic units (facts) that were labeled as \emph{supported} (S) or \emph{not-supported} (NS) by human annotators. We assume that this annotation is the ground truth.

The AskH, ELI5, Books and LFObj datasets are unlabeled and consist of collections of prompts. Specifically, the AskH and ELI5 datasets each contain 200 questions sourced from the Reddit forums r/AskHistorians and r/explainlikeimfive, respectively. The Books dataset comprises 200 paragraphs, sampled as 10 excerpts from each of 20 non-fiction books published between 2023 and 2024. Our version of the LFObj dataset is a curated subset of the original collection \cite{wei2024longform}, consisting of 10 prompts randomly selected from those related to objects spanning 38 distinct topics. For each prompt across these datasets, we generated a long-form response -- up to two paragraphs in length -- using the llama-3.3-70b-instruct model \cite{touvron2023llama}.

Additionally, we constructed a new dataset, Conflicts, comprising 1,000 claims (or atomic units) randomly sampled from the recent ConflictBank benchmark \cite{su2024conflictbank}. Each claim, originally extracted from Wikidata, is considered true (i.e., supported). For every claim, we include two associated contexts: one supporting context (the default in ConflictBank) and one conflicting context (representing misinformation, also provided by ConflictBank). Notably, these two contexts are mutually contradictory, thus offering a controlled setting for our long-form factuality evaluation.

\paragraph{Measures of Performance.} 
For each dataset $\mathcal{D}$ and each competing assessor, we report the factual precision ($Pr$) and the $F_1@K$ score, averaged over all prompts in $\mathcal{D}$. If $\mathcal{D}$ includes annotated atomic units (i.e., ground truth labels), we additionally report the standard $F_1$ score and the mean absolute error (MAE), defined as:
\begin{equation}
    \textrm{MAE} = \frac{1}{|\cD|}\sum_{j=1}^{|\cD|} |Pr_j - Pr^*_j|
\end{equation}
\noindent where $Pr_j$ is the predicted precision and $Pr^*_j$ is the ground-truth factual precision for the $j$-th instance. Since the \textsc{FactReasoner} assessors calculate the posterior marginal distributions of the atoms, we also compute the $\cE$-measure. Finally, we include the mean number of atoms classified as supported (\#S), contradicted (\#C), and undecided (\#U).

\paragraph{External Knowledge Sources.} We consider two external knowledge sources: Wikipedia and Google Search results. For a given atom, the top $k$ results are retrieved as contexts either from wikipedia.org using the Wikipedia retriever available from LangChain\footnote{\texttt{https://python.langchain.com}}, or from google.com using the Serper API\footnote{\texttt{https://serper.dev}}. In both cases, a context is a tuple $(t,l,s,d)$, where $t$ is the title of the wiki/web-page, $l$ is the link, $s$ is a short text snippet or summary and $d$ is the content retrieved from $l$ (but capped at max 4000 characters). We used $k=3$ for the Wikipedia retriever and $k=5$ for the Google Search results \cite{factscore2023emnlp,wei2024longform}.

To ensure consistent evaluation across all datasets, we decompose each generated response into its constituent atomic units and revise them using the same llama-3.3-70b-instruct model. Additionally, we retrieve and cache the relevant contextual information for each atom from the two designated knowledge sources. This standardized setup allows all competing assessors to be evaluated on an identical set of atoms and associated contexts. Table~\ref{tab:datasets} summarizes the key properties of the datasets, including the number of prompts, the mean number of atoms per response, and the median number of atoms ($K$), which is used in computing the $F_1@K$ metric. For the labeled datasets, we also report the true number of supported atoms ($S^*$) and ground-truth precision (Pr$^*$).

\begin{table}[t!]
    \centering
    \resizebox{\linewidth}{!}{%
    \begin{tabular}{l|r|r|r|r|r|r|r|r}
       Assessor & \# S & \# C & \# U & Pr$\uparrow$ & $F_1$ $\uparrow$ & $F_1@K$ $\uparrow$ & MAE$\downarrow$ & $\cE$ $\downarrow$ \\
       \toprule
       \multicolumn{9}{c}{BERT-based relation model: \texttt{albert-xlarge-vitaminc-mnli}}\\
       \midrule
       FR1  & 12 &  5 & 12 & 0.40 & 0.66 & 0.39 & 0.25 & 0.11 \\
       FR2  & 10 & 16 &  4 & 0.32 & 0.53 & 0.31 & 0.34 & 0.09 \\
       FR3  & 10 & 16 &  4 & 0.32 & 0.53 & 0.31 & 0.33 & 0.09 \\       
       \toprule
       \multicolumn{9}{c}{LLM-based relation model: \texttt{llama-3.1-70b-instruct}}\\
       \midrule
       FR1  & 13 & 1 & 16  & 0.41 & 0.70 & 0.41 & 0.23 & 0.10 \\
       FR2  & {\bf 19} & 2 &  9  & {\bf 0.60} & {\bf 0.83} & {\bf 0.59} & {\bf 0.11} & {\bf 0.06} \\
       FR3  & {\bf 19} & 2 &  9  & {\bf 0.60} & {\bf 0.83} & {\bf 0.59} & {\bf 0.11} & {\bf 0.06} \\
    \end{tabular}}
    \caption{Results for the \texttt{vitc}- and \texttt{llama}-based relation models used by FactReasoner's Evaluator stage.}
    \label{tab:nli}
\end{table}

\subsection{Evaluating the Relation Model}
We first evaluate the relation model used by the Evaluator stage of the FactReasoner assessor to extract the atom-context and context-context relationships required to construct the graphical model. Specifically, we consider two relation models based on a standard BERT-based model such as \texttt{vitc} \cite{schuster2021vitc} and on a larger LLM such as \texttt{llama-3.1-70b-instruct} \cite{touvron2023llama} with a suitable few-shots prompt. 

Table \ref{tab:nli} shows the results obtained for the FR1, FR2 and FR3 assessors employing the two types of relation models on the Biographies dataset using Wikipedia retrieved contexts. We observe that using the LLM-based relation model which predicts entailments much more accurately than the BERT-based one leads to significant improvements in performance, especially for the FR2 and FR3 variants. For example, the \texttt{llama}-based FR2 achieves an $F_1$ score nearly twice as high compared with the \texttt{vitc}-based one (i.e., 0.83 versus 0.53). For this reason, we only employ LLM-based relation models for now on (see also the Appendix for more details).

\begin{figure}[t!]
    \centering
    \includegraphics[width=\linewidth]{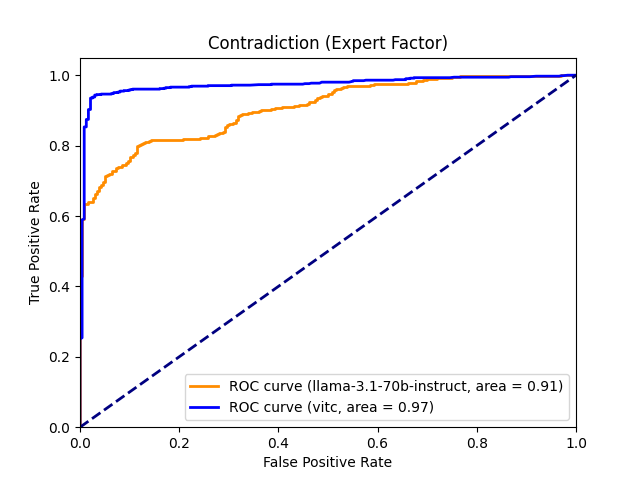}
    \includegraphics[width=\linewidth]{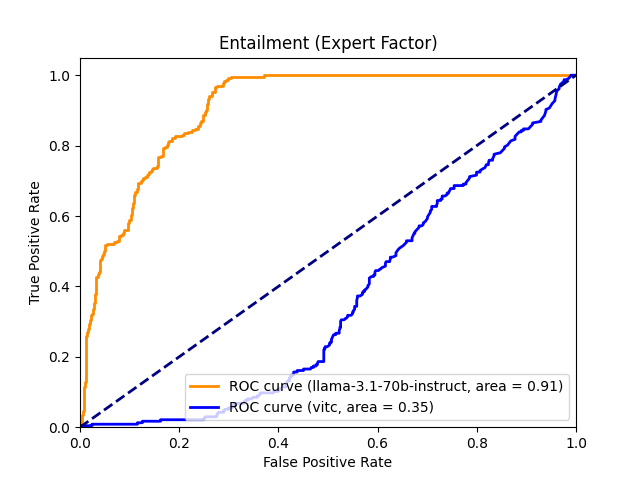}
    \caption{ROC curves for the \texttt{vitc}- and \texttt{llama}-based relation models predicting contradiction and entailment.}
    \label{fig:nli-llama2}
\end{figure}

\begin{figure}[t!]
    \centering
    \includegraphics[width=\linewidth]{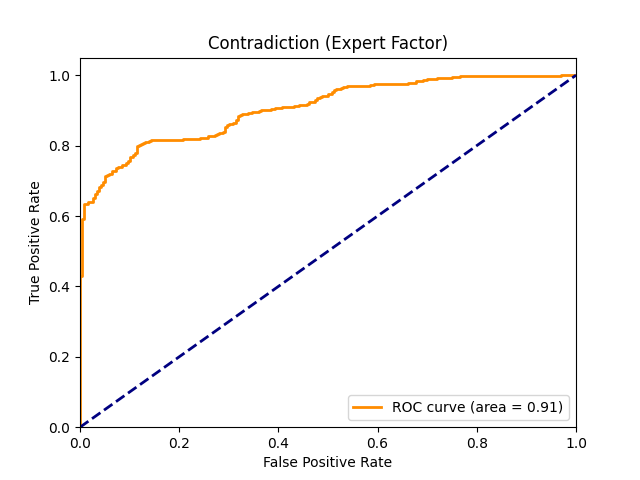}
    \includegraphics[width=\linewidth]{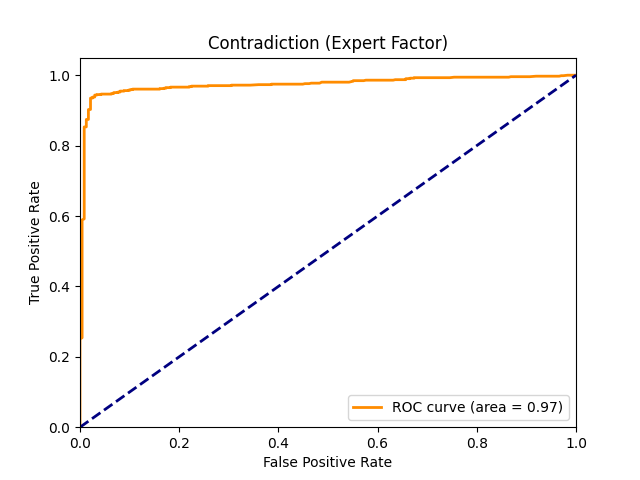}
    \caption{ROC curves for the \texttt{llama}- (top) \texttt{vitc}-based (bottom) relation models predicting contradiction on the Expert FACTOR dataset.}
    \label{fig:nli-contr-expert}
\end{figure}
\begin{figure}[t!]
    \centering
    \includegraphics[width=\linewidth]{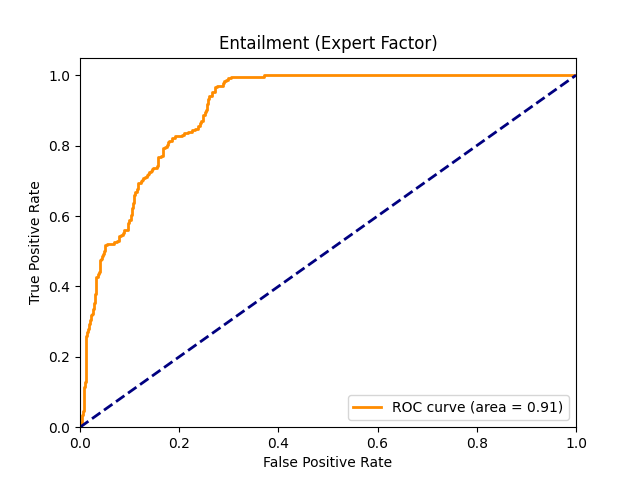}
    \includegraphics[width=\linewidth]{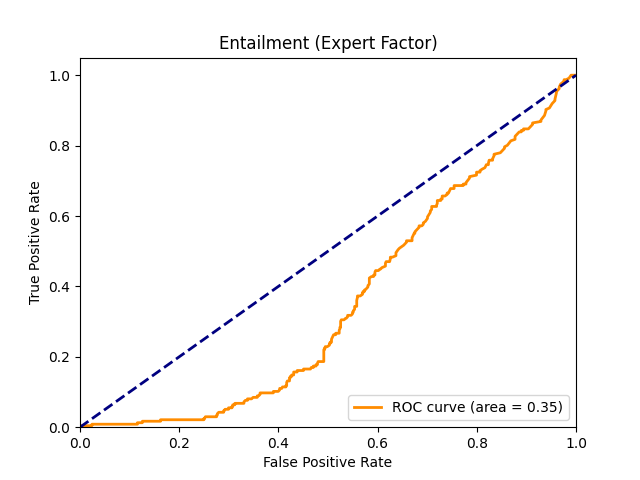}
    \caption{ROC curves for the \texttt{llama}- (top) \texttt{vitc}-based (bottom) relation models predicting entailment on the Expert FACTOR dataset.}
    \label{fig:nli-entail-expert}
\end{figure}
\begin{figure}[t!]
    \centering
    \includegraphics[width=\linewidth]{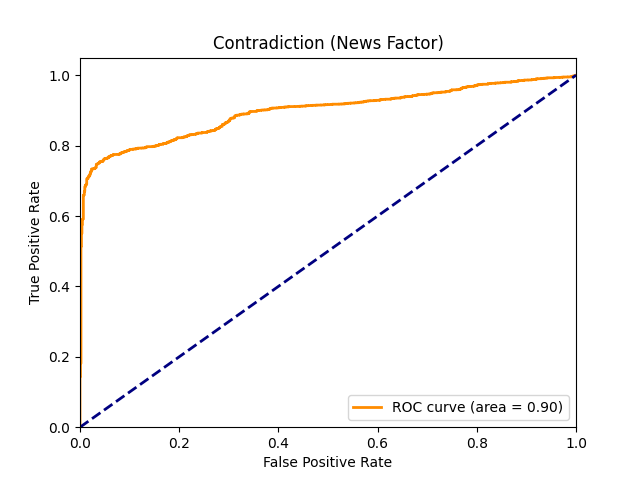}
    \includegraphics[width=\linewidth]{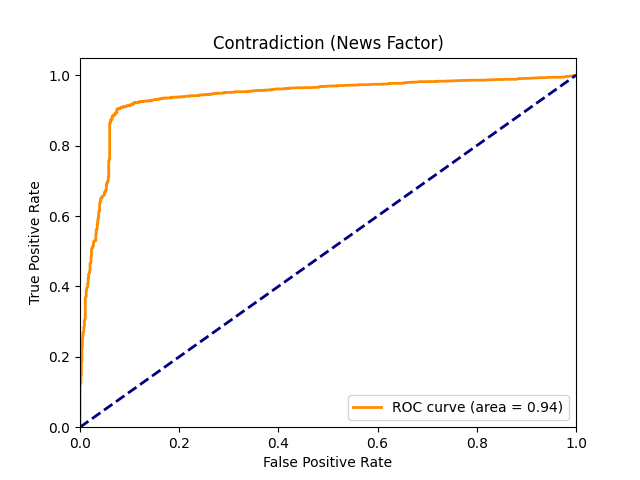}
    \caption{ROC curves for the \texttt{llama}- (top) \texttt{vitc}-based (bottom) relation models predicting contradiction on the News FACTOR dataset.}
    \label{fig:nli-contr-news}
\end{figure}
\begin{figure}[t!]
    \centering
    \includegraphics[width=\linewidth]{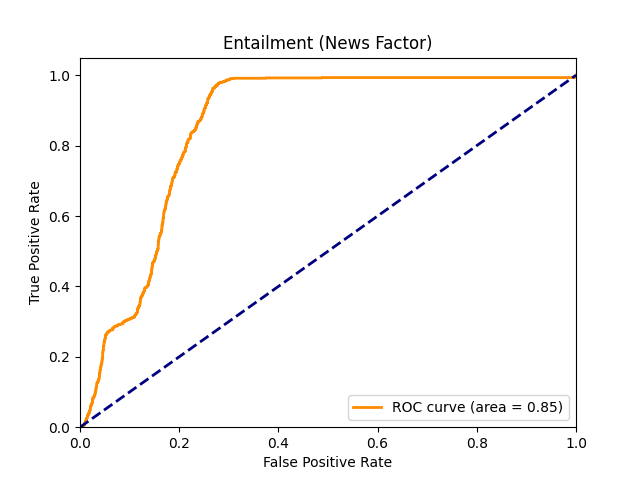}
    \includegraphics[width=\linewidth]{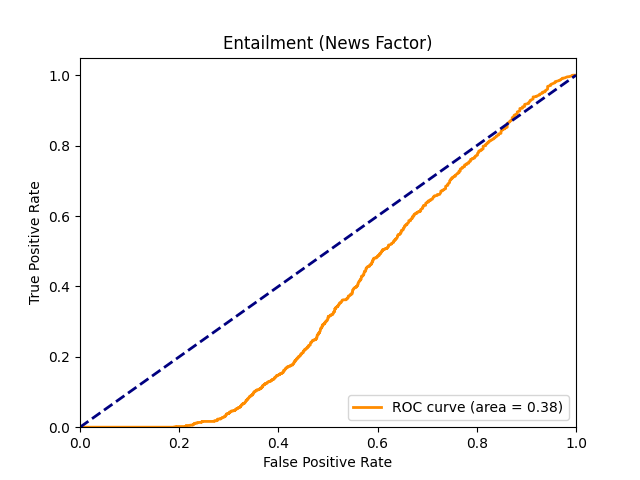}
    \caption{ROC curves for the \texttt{llama}- (top) \texttt{vitc}-based (bottom) relation models predicting entailment on the News FACTOR dataset.}
    \label{fig:nli-entail-news}
\end{figure}

Figure \ref{fig:nli-llama2} plots the ROC curves for predicting contradiction and entailment relationships on the Expert FACTOR dataset \cite{factor2024ecacl}. We see that the \texttt{vitc}-based model predicts contradictions fairly accurately compared with the \texttt{llama}-based one, but performs rather poorly on predicting the entailment relations. 

Figures \ref{fig:nli-contr-expert} and \ref{fig:nli-entail-expert} plot the ROC curves for predicting the contradiction and entailment relationships by the \texttt{llama}- and \texttt{vitc}-based relation models on the Expert FACTOR dataset \cite{factor2024ecacl}. Figures \ref{fig:nli-contr-news} and \ref{fig:nli-entail-news} plot the ROC curves for predicting the contradiction and entailment relationships by the same relation models on the News FACTOR dataset \cite{factor2024ecacl}

\subsection{Calibration Results}
We confirm that the predictions of \textsc{FactReasoner} are well calibrated. For example, on the labeled Biographies dataset with Wikipedia contexts, the mean Brier score for FR2 using the llama-3.1-70b-instruct model is 0.18 ($\pm$0.10), which clearly indicates a reasonably good calibration (perfect calibration corresponds to a Brier score of 0). Unfortunately, it is not possible to calculate Brier scores for the prompt-based methods because they do not compute probabilities associated with the atoms.

\subsection{Additional Results on Labeled and Unlabeled Datasets}
In Table \ref{tab:bio-google} we show the results obtained on the same Biographies dataset but using Google Search results as contexts. We observe a similar pattern of the results compared with the previous case, namely FV and VS being more conservative than the FR assessors. However, we notice that in this case there are many more atoms labeled as supported (\#S) and consequently more false positives which is reflected in the slightly higher MAE values for all competing assessors. We believe that this is most likely caused by the slightly noisier contexts compared with the Wikipedia only based ones which eventually leads to more spurious entailment relationships than in the previous case. As before, we note that the relatively simple prompt employed by FS leads to large numbers of atoms labeled as supported.

Tables \ref{tab:bio-wikipedia2} and \ref{tab:bio-google2} contain the detailed results obtained on the labeled Biographies dataset including the standard deviations for each of the reported performance measures.

Tables \ref{tab:askh-wikipedia2}, \ref{tab:books-wikipedia2}, \ref{tab:eli5-wikipedia2} and \ref{tab:lfobj-wikipedia2} report the detailed results obtained on the unlabeled datasets AskH, Books, ELI5 and LFObj using Wikipedia retrieved contexts. Tables \ref{tab:askh-google2}, \ref{tab:books-google2}, \ref{tab:eli5-google2} and \ref{tab:lfobj-google2} show the detailed results obtained on the unlabeled datasets Books, ELI5 and LFObj using Google Search results based contexts. All these additional results show a similar pattern to those reported for the AskH dataset in the main paper.

\begin{table}[t!]
    \centering
    \resizebox{\linewidth}{!}{%
    \begin{tabular}{l|r|r|r|r|r|r|r|r}
       Method & \# S & \# C & \# U & Pr & $F_1$ & $F_1@K$ & MAE & $\cE$ \\
       \toprule
       \multicolumn{9}{c}{\texttt{granite-3.0-8b-instruct}}\\
       \midrule
       FS   & 24 & 6  &    & 0.76  & 0.80   & 0.73  & 0.15 &   \\
       FV   & 20 & 2  & 8  & 0.64  & 0.74   & 0.62  & 0.14 &  \\
       VS   & 21 & 1  & 8  & 0.67  & 0.74   & 0.65  & 0.14 &  \\
       FR1  & 23 & 3  & 4  & 0.73  & 0.79   & 0.70  & 0.14 & 0.08 \\
       FR2  & 24 & 5  & 0  & 0.78 & 0.80  & 0.75  & 0.19 & 0.04 \\
       FR3  & 24 & 5  & 0  & 0.78 & 0.79  & 0.74  & 0.18 & 0.04 \\
       \midrule
       \multicolumn{9}{c}{\texttt{llama-3.1-70b-instruct}}\\
       \midrule
       FS   & 23 & 7  &   & 0.73  &  0.82  & 0.71  & 0.14 &  \\
       FV   & 23 & 3  & 4  & 0.72  & 0.82   & 0.70   & 0.13 &  \\
       VS   & 23 & 1  & 6 & 0.72  & 0.81   & 0.70  & 0.13 &  \\
       FR1  & 21 & 2  & 7 & 0.66  & 0.81  & 0.64  & 0.11 & 0.06 \\
       FR2  & 24 & 2 & 3  & 0.77  & 0.83  & 0.74  & 0.16 & 0.03 \\
       FR3  & 24 & 2 & 3  & 0.77  & 0.83  & 0.74  & 0.16 & 0.03 \\
       \midrule
       \multicolumn{9}{c}{\texttt{mixtral-8x22b-instruct}}\\
       \midrule
       FS   & 24 & 6  &   & 0.75 & 0.83  & 0.72  & 0.15 &  \\
       FV   & 22 & 2  & 5  & 0.71  & 0.82  & 0.69 & 0.12 &  \\
       VS   & 23 & 1  & 5  & 0.73  & 0.81  & 0.71 & 0.13 &  \\
       FR1  & 22 & 1  & 6  & 0.71  & 0.81  & 0.69 & 0.13 & 0.05 \\
       FR2  & 25 & 1 & 3  & 0.81  & 0.82  & 0.77  & 0.19 & 0.03 \\
       FR3  & 25 & 2 & 3  & 0.80  & 0.82  & 0.77  & 0.19 & 0.03 \\

    \end{tabular}}
    \caption{Results obtained on the labeled Biographies dataset using Google Search retrieved contexts.}
    \label{tab:bio-google}
\end{table}

\begin{table}[t!]
    \centering
    \resizebox{\linewidth}{!}{%
    \begin{tabular}{l|r|r|r|r|r|r|r|r}
       Assessor & \# S & \# C & \# U & Pr & $F_1$ & $F_1@K$ & MAE & $\cE$ \\
       \toprule
       \multicolumn{9}{c}{\texttt{granite-3.0-8b-instruct}}\\
       \midrule
       FS   & 18$\pm$8 & 12$\pm$5  &     & 0.59$\pm$0.17  & 0.70$\pm$0.17  & 0.57$\pm$0.20  & 0.17$\pm$0.14 &   \\
       FV   & 14$\pm$7 & 2$\pm$1  & 14$\pm$6  & 0.45$\pm$0.19  & 0.67$\pm$0.15  & 0.44$\pm$0.21  & 0.21$\pm$0.14 &  \\
       VS   &  15$\pm$8  & 8$\pm$4  &  6$\pm$3  & 0.49$\pm$0.20  &  0.64$\pm$0.19 &  0.48$\pm$0.22  &  0.21$\pm$0.14   &  \\
       FR1  & 14$\pm$6 & 2$\pm$2 & 14$\pm$6  & 0.43$\pm$0.20 & 0.70$\pm$0.15  & 0.43$\pm$0.21 & 0.22$\pm$0.13  &  0.12$\pm$0.01 \\
       FR2  & 20$\pm$6 & 4$\pm$3 &  6$\pm$3  & {\bf 0.62$\pm$0.21} & {\bf 0.78$\pm$0.15}  & {\bf 0.61$\pm$0.23} & {\bf 0.12$\pm$0.13}  & 0.06$\pm$0.01 \\
       FR3  & 19$\pm$6 & 4$\pm$3 &  6$\pm$3  & 0.60$\pm$0.19 & 0.78$\pm$0.14  & 0.59$\pm$0.22 & 0.13$\pm$0.13  &  0.06$\pm$0.01 \\
       \midrule
       \multicolumn{9}{c}{\texttt{llama-3.1-70b-instruct}}\\
       \midrule
       FS   & 19$\pm$8 & 12$\pm$5  &     & 0.59$\pm$0.20  & 0.73$\pm$0.16  & 0.58$\pm$0.20  & 0.16$\pm$0.14 &  \\
       FV   & 15$\pm$8 & 1$\pm$1  & 14$\pm$6  & 0.47$\pm$0.20  & 0.73$\pm$0.15  & 0.47$\pm$0.22  & 0.19$\pm$0.12 &  \\
       VS   & 12$\pm$8   & 0  &  18$\pm$7 & 0.38$\pm$0.21  &  0.64$\pm$0.18 &  0.38$\pm$0.23  &  0.27$\pm$0.15 &  \\
       FR1  & 13$\pm$8 & 1$\pm$2 & 16$\pm$6  & 0.42$\pm$0.20  & 0.71$\pm$0.15  & 0.42$\pm$0.21  & 0.23$\pm$0.13 & 0.10$\pm$0.02 \\
       FR2  & 19$\pm$9 & 2$\pm$2 &  9$\pm$5  & {\bf 0.60$\pm$0.20}  & {\bf 0.83$\pm$0.13}  & {\bf 0.59$\pm$0.24}  & {\bf 0.11$\pm$0.11} & {\bf 0.06$\pm$0.02} \\
       FR3  & 19$\pm$9 & 2$\pm$2 &  9$\pm$5  & {\bf 0.60$\pm$0.20}  & {\bf 0.83$\pm$0.14}  & {\bf 0.59$\pm$0.24}  & {\bf 0.11$\pm$0.12} & {\bf 0.06$\pm$0.02} \\
       \midrule
       \multicolumn{9}{c}{\texttt{mixtral-8x22b-instruct}}\\
       \midrule
       FS   & 19$\pm$8 & 12$\pm$5 &    & 0.59$\pm$0.18  & 0.74$\pm$0.16  & 0.58$\pm$0.20  & 0.16$\pm$0.13 &  \\
       FV   & 15$\pm$7 & 1$\pm$1 & 13$\pm$5  & 0.49$\pm$0.18  & 0.72$\pm$0.14  & 0.48$\pm$0.21  & 0.19$\pm$0.12 &  \\
       VS   & 13$\pm$7  & 1$\pm$1 & 15$\pm$6  & 0.42$\pm$0.18  & 0.65$\pm$0.16 &  0.42$\pm$0.20  &  0.25$\pm$0.14    &  \\
       FR1  & 14$\pm$8 & 0$\pm$1 & 15$\pm$6 & 0.44$\pm$0.20  & 0.72$\pm$0.15  & 0.44$\pm$0.22  & 0.21$\pm$0.13  & 0.10$\pm$0.02 \\
       FR2  & 20$\pm$9 & 1$\pm$1 &  8$\pm$5 & {\bf 0.63$\pm$0.20}  & {\bf 0.83$\pm$0.14}  & {\bf 0.62$\pm$0.24}  & {\bf 0.11$\pm$0.11}  & {\bf 0.07$\pm$0.01} \\
       FR3  & 20$\pm$9 & 1$\pm$1 &  9$\pm$5 & {\bf 0.64$\pm$0.21}  & {\bf 0.83$\pm$0.14}  & {\bf 0.62$\pm$0.24}  & {\bf 0.11$\pm$0.12}  & {\bf 0.07$\pm$0.01} \\

    \end{tabular}}
    \caption{Results obtained on the labeled Biographies dataset using Wikipedia retrieved contexts.}
    \label{tab:bio-wikipedia2}
\end{table}

\begin{table}[t!]
    \centering
    \resizebox{\linewidth}{!}{%
    \begin{tabular}{l|r|r|r|r|r|r|r|r}
       Method & \# S & \# C & \# U & Pr & $F_1$ & $F_1@K$ & MAE & $\cE$ \\
       \toprule
       \multicolumn{9}{c}{\texttt{granite-3.0-8b-instruct}}\\
       \midrule
       FS   & 24$\pm$10 & 6$\pm$5  &   & 0.76$\pm$0.20  & 0.80$\pm$0.17   & 0.73$\pm$0.23  & 0.15$\pm$0.14 &   \\
       FV   & 20$\pm$8 & 2$\pm$2  & 8$\pm$4  & 0.64$\pm$0.18  & 0.74$\pm$0.16   & 0.62$\pm$0.21  & 0.14$\pm$0.12 &  \\
       VS   & 21$\pm$9 & 1$\pm$1  & 8$\pm$4  & 0.67$\pm$0.18  & 0.74$\pm$0.17   & 0.65$\pm$0.21  & 0.14$\pm$0.12 &  \\
       FR1  & 23$\pm$9  &  3$\pm$2 & 4$\pm$3  & 0.73$\pm$0.19  & 0.79$\pm$0.15  & 0.70$\pm$0.22  & 0.14$\pm$0.14 & 0.08$\pm$0.01 \\
       FR2  & 24$\pm$10 & 5$\pm$5 & 0$\pm$1  & 0.78$\pm$0.20 & 0.80$\pm$0.18  & 0.75$\pm$0.23  & 0.19$\pm$0.16 & 0.04$\pm$0.01 \\
       FR3  & 24$\pm$10 & 5$\pm$6 & 0$\pm$1  & 0.78$\pm$0.21 & 0.79$\pm$0.18  & 0.74$\pm$0.24  & 0.18$\pm$0.16 & 0.04$\pm$0.01 \\
       \midrule
       \multicolumn{9}{c}{\texttt{llama-3.1-70b-instruct}}\\
       \midrule
       FS   & 23$\pm$10 & 7$\pm$5  &   & 0.73$\pm$0.20  &  0.82$\pm$0.15  & 0.71$\pm$0.23  & 0.14$\pm$0.13 &  \\
       FV   & 23$\pm$10 & 3$\pm$2 & 4$\pm$3  & 0.72$\pm$0.20  & 0.82$\pm$0.16   & 0.70$\pm$0.23  & 0.13$\pm$0.12 &  \\
       VS   & 23$\pm$10 & 1$\pm$1  & 6$\pm$5 & 0.72$\pm$0.21  & 0.81$\pm$0.15   & 0.70$\pm$0.24  & 0.13$\pm$0.12 &  \\
       FR1  & 21$\pm$9 & 2$\pm$1 & 7$\pm$4  & 0.66$\pm$0.22  & 0.81$\pm$0.15  & 0.64$\pm$0.22  & 0.11$\pm$0.12 & 0.06$\pm$0.01 \\
       FR2  & 24$\pm$10 & 2$\pm$2 & 3$\pm$3  & 0.77$\pm$0.20  & 0.83$\pm$0.17  & 0.74$\pm$0.23  & 0.16$\pm$0.14 & 0.03$\pm$0.01 \\
       FR3  & 24$\pm$10 & 2$\pm$2 & 3$\pm$3  & 0.77$\pm$0.20  & 0.83$\pm$0.17  & 0.74$\pm$0.23  & 0.16$\pm$0.14 & 0.03$\pm$0.01 \\
       \midrule
       \multicolumn{9}{c}{\texttt{mixtral-8x22b-instruct}}\\
       \midrule
       FS   & 24$\pm$10 & 6$\pm$5  &   & 0.75$\pm$0.20  & 0.83$\pm$0.16  & 0.72$\pm$0.23  & 0.15$\pm$0.14 &  \\
       FV   & 22$\pm$9 & 2$\pm$2  & 5$\pm$4  & 0.71$\pm$0.20  & 0.82$\pm$0.15  & 0.69$\pm$0.23  & 0.12$\pm$0.12 &  \\
       VS   & 23$\pm$10 & 1$\pm$1  & 5$\pm$4  & 0.73$\pm$0.21  & 0.81$\pm$0.16  & 0.71$\pm$0.24  & 0.13$\pm$0.13 &  \\
       FR1  & 22$\pm$9 & 1$\pm$1 & 6$\pm$4  & 0.71$\pm$0.20  & 0.81$\pm$0.15  & 0.69$\pm$0.23  & 0.13$\pm$0.13 & 0.05$\pm$0.01 \\
       FR2  & 25$\pm$10 & 1$\pm$2 & 3$\pm$3  & 0.81$\pm$0.18  & 0.82$\pm$0.17  & 0.77$\pm$0.22  & 0.19$\pm$0.16 & 0.03$\pm$0.01 \\
       FR3  & 25$\pm$10 & 2$\pm$4 & 3$\pm$3  & 0.80$\pm$0.19  & 0.82$\pm$0.17  & 0.77$\pm$0.22  & 0.19$\pm$0.17 & 0.03$\pm$0.01 \\

    \end{tabular}}
    \caption{Results obtained on the labeled Biographies dataset using Google Search retrieved contexts.}
    \label{tab:bio-google2}
\end{table}

\begin{table}[t!]
    \centering
    \resizebox{\linewidth}{!}{%
    \begin{tabular}{l|r|r|r|r|r|r}
       Assessor & \# S & \# C & \# U & Pr $\uparrow$ & $F_1@K$ $\uparrow$ & $\cE$ $\downarrow$ \\
       \toprule
       \multicolumn{7}{c}{\texttt{granite-3.0-8b-instruct}}\\
       \midrule
       FS   & 17$\pm$6 & 5$\pm$3  &   & 0.76$\pm$0.15  & 0.74$\pm$0.17  &    \\
       FB   &  8$\pm$4 & 0$\pm$1  & 13$\pm$4  & 0.35$\pm$0.15  & 0.36$\pm$0.17  &    \\
       FV   & 12$\pm$5 & 4$\pm$2  &  5$\pm$2 & 0.55$\pm$0.16  & 0.55$\pm$0.18   &    \\
       FR1 (ours)  &  4$\pm$3 & 1$\pm$1 & 16$\pm$4 & 0.19$\pm$0.14  & 0.19$\pm$0.13  & 0.14$\pm$0.01  \\
       FR2 (ours) & 10$\pm$6 & 9$\pm$5 & 2$\pm$2  & 0.46$\pm$0.22  & 0.47$\pm$0.24  & 0.09$\pm$0.03  \\
       FR3 (ours) & 11$\pm$6 & 8$\pm$4 & 2$\pm$2  & 0.47$\pm$0.22  & 0.48$\pm$0.24  & 0.10$\pm$0.03   \\
       \midrule
       \multicolumn{7}{c}{\texttt{llama-3.1-70b-instruct}}\\
       \midrule
       FS   &  15$\pm$5  & 7$\pm$3  &  & 0.69$\pm$0.15  & 0.68$\pm$0.16  &    \\
       FB   &   8$\pm$5 &  0 & 13$\pm$4  & 0.37$\pm$0.19  & 0.38$\pm$0.21   &  \\
       FV   &   5$\pm$4 &  0 & 16$\pm$5  & 0.25$\pm$0.16  & 0.25$\pm$0.18   &  \\
       FR1 (ours) &  5$\pm$4 & 0  & 17$\pm$4  & 0.21$\pm$0.16  & 0.22$\pm$0.18  & 0.13$\pm$0.02 \\
       FR2 (ours) & 10$\pm$7 & 1$\pm$1  & 10$\pm$5  & 0.45$\pm$0.26  & 0.46$\pm$0.28  & 0.09$\pm$0.04 \\
       FR3 (ours)  & 10$\pm$7 & 1$\pm$1  & 10$\pm$5  & 0.44$\pm$0.25  & 0.45$\pm$0.27  & 0.09$\pm$0.04 \\
       \midrule
       \multicolumn{7}{c}{\texttt{mixtral-8x22b-instruct}}\\
       \midrule
       FS   & 16$\pm$5  & 6$\pm$3  &   &  0.71$\pm$0.15  & 0.70$\pm$0.16  &  \\
       FB   &  9$\pm$5 & 0  & 12$\pm$4  & 0.43$\pm$0.17  & 0.43$\pm$0.19  &  \\
       FV   &  7$\pm$4 & 0  & 14$\pm$5  & 0.34$\pm$0.17  & 0.34$\pm$0.19  &  \\
       FR1 (ours) &  5$\pm$4 & 0  & 17$\pm$4  & 0.22$\pm$0.18  & 0.23$\pm$0.20  & 0.12$\pm$0.02 \\
       FR2 (ours) & 11$\pm$5 & 0  & 11$\pm$5  & 0.46$\pm$0.28  & 0.47$\pm$0.30  & 0.09$\pm$0.04 \\
       FR3 (ours) & 11$\pm$8 & 0  & 11$\pm$5  & 0.46$\pm$0.30  & 0.47$\pm$0.30  & 0.09$\pm$0.04 \\

    \end{tabular}}
    \caption{Results obtained on the unlabeled AskH dataset using Wikipedia retrieved contexts.}
    \label{tab:askh-wikipedia2}
\end{table}

\begin{table}[t!]
    \centering
    \resizebox{\linewidth}{!}{%
    \begin{tabular}{l|r|r|r|r|r|r}
       Assessor & \# S & \# C & \# U & Pr $\uparrow$ & $F_1@K$ $\uparrow$ & $\cE$ $\downarrow$ \\
       \toprule
       \multicolumn{7}{c}{\texttt{granite-3.0-8b-instruct}}\\
       \midrule
       FS   & 18$\pm$6  & 3$\pm$2  &   & 0.82$\pm$0.13  & 0.81$\pm$0.15  &     \\
       FV   &  14$\pm$5 & 1$\pm$1  & 7$\pm$3  & 0.62$\pm$0.16  & 0.62$\pm$0.19   &    \\
       VS   &  14$\pm$5 & 3$\pm$2  & 3$\pm$2  & 0.65$\pm$0.15  & 0.65$\pm$0.15   &    \\
       FR1 (ours)  & 13$\pm$5 & 4$\pm$2  & 4$\pm$2  & 0.60$\pm$0.17  & 0.60$\pm$0.20  &  0.08$\pm$0.02  \\
       FR2 (ours) & 14$\pm$8 & 7$\pm$5  & 0  & 0.63$\pm$0.27  & 0.62$\pm$0.28  &  0.04$\pm$0.03  \\
       FR3 (ours) & 15$\pm$7 & 7$\pm$5  & 0  & 0.67$\pm$0.24  & 0.66$\pm$0.25  &  0.06$\pm$0.03  \\
       \midrule
       \multicolumn{7}{c}{\texttt{llama-3.1-70b-instruct}}\\
       \midrule
       FS   & 18$\pm$5 & 3$\pm$2  &   & 0.82$\pm$0.12  & 0.80$\pm$0.14  &  \\
       FV   & 16$\pm$6 & 1$\pm$1  & 5$\pm$3  & 0.71$\pm$0.18  & 0.70$\pm$0.20   &  \\
       VS   & 15$\pm$6 & 0  & 7$\pm$3  & 0.66$\pm$0.18  & 0.65$\pm$0.20   &  \\
       FR1 (ours) & 12$\pm$6 & 1$\pm$1  & 8$\pm$4  & 0.53$\pm$0.19  & 0.54$\pm$0.22  & 0.08$\pm$0.03 \\
       FR2 (ours) & 17$\pm$6 & 1$\pm$1  & 3$\pm$3  & 0.76$\pm$0.18  & 0.74$\pm$0.20  & 0.04$\pm$0.03 \\
       FR3 (ours) & 17$\pm$6 & 2$\pm$1  & 3$\pm$3  & 0.75$\pm$0.18  & 0.74$\pm$0.20  & 0.04$\pm$0.03 \\
       \midrule
       \multicolumn{7}{c}{\texttt{mixtral-8x22b-instruct}}\\
       \midrule
       FS   & 18$\pm$6 & 3$\pm$2  &   & 0.82$\pm$0.13   & 0.80$\pm$0.15  &  \\
       FV   & 15$\pm$6 & 0$\pm$1  & 6$\pm$3  &  0.67$\pm$0.18  & 0.67$\pm$0.21  &  \\
       VS   & 15$\pm$6 & 0$\pm$1  & 6$\pm$3  &  0.68$\pm$0.18  & 0.67$\pm$0.20  &  \\
       FR1 (ours) & 14$\pm$6 & 0  & 8$\pm$4  & 0.60$\pm$0.20  & 0.60$\pm$0.22  & 0.07$\pm$0.03 \\
       FR2 (ours) & 18$\pm$7 & 0  & 3$\pm$3  & 0.80$\pm$0.17  & 0.79$\pm$0.19  & 0.04$\pm$0.03 \\
       FR3 (ours) & 18$\pm$7 & 0  & 3$\pm$3  & 0.80$\pm$0.17  & 0.79$\pm$0.19  & 0.04$\pm$0.03 \\

    \end{tabular}}
    \caption{Results obtained on the unlabeled AskH dataset using Google Search retrieved contexts.}
    \label{tab:askh-google2}
\end{table}

\begin{table}[t!]
    \centering
    \resizebox{\linewidth}{!}{%
    \begin{tabular}{l|r|r|r|r|r|r}
       Method & \# S & \# C & \# U & Pr & $F_1@K$ & $\cE$ \\
       \toprule
       \multicolumn{7}{c}{\texttt{granite-3.0-8b-instruct}}\\
       \midrule
       FS   & 17$\pm$6 & 6$\pm$4  &   & 0.72$\pm$0.19  & 0.71$\pm$20   &    \\
       FV   & 9$\pm$5 & 0  & 13$\pm$5  & 0.38$\pm$0.18  & 0.38$\pm$0.18   &    \\
       VS   & 15$\pm$6 & 3$\pm$2  & 4$\pm$2  & 0.63$\pm$0.16  & 0.63$\pm$0.18  &    \\
       FR1  & 8$\pm$6 & 0$\pm$0  & 14$\pm$6  & 0.34$\pm$0.23  & 0.34$\pm$0.24  & 0.11$\pm$0.03   \\
       FR2  & 15$\pm$9 & 0$\pm$1  & 7$\pm$6  & 0.64$\pm$0.29  & 0.63$\pm$0.29  & 0.06$\pm$0.04   \\
       FR3  & 13$\pm$8 & 7$\pm$6  & 2$\pm$3  & 0.55$\pm$0.27  & 0.54$\pm$0.28  & 0.09$\pm$0.03   \\
       \midrule
       \multicolumn{7}{c}{\texttt{llama-3.1-70b-instruct}}\\
       \midrule
       FS   & 16$\pm$6 & 7$\pm$4  & & 0.69$\pm$0.18  & 0.68$\pm$0.19   &  \\
       FV   & 10$\pm$6 & 0  & 12$\pm$5  & 0.43$\pm$0.22  & 0.43$\pm$0.23   &  \\
       VS   &  5$\pm$4 & 0  & 17$\pm$4  & 0.24$\pm$0.18  & 0.24$\pm$0.18   &  \\
       FR1  & 5$\pm$5 & 0$\pm$0  & 17$\pm$6  & 0.24$\pm$0.18  & 0.24$\pm$0.19  & 0.12$\pm$0.02 \\
       FR2  & 11$\pm$8 & 1$\pm$1  & 10$\pm$6  & 0.49$\pm$0.29  & 0.49$\pm$0.29  & 0.09$\pm$0.04 \\
       FR3  & 12$\pm$8 & 2$\pm$2  & 9$\pm$6  & 0.49$\pm$0.28  & 0.50$\pm$0.29  & 0.09$\pm$0.04 \\
       \midrule
       \multicolumn{7}{c}{\texttt{mixtral-8x22b-instruct}}\\
       \midrule
       FS   & 17$\pm$6 & 6$\pm$4  &   &  0.72$\pm$0.19  & 0.71$\pm$0.20  &  \\
       FV   & 11$\pm$6 & 0  & 11$\pm$5  & 0.50$\pm$0.21   & 0.50$\pm$0.21  &  \\
       VS   & 10$\pm$6 & 0  & 12$\pm$6  & 0.43$\pm$0.22   & 0.43$\pm$0.22  &  \\
       FR1  & 6$\pm$5 & 0$\pm$0  & 17$\pm$6  & 0.25$\pm$0.20  & 0.25$\pm$0.21  & 0.12$\pm$0.03 \\
       FR2  & 12$\pm$8 & 0$\pm$0  & 10$\pm$6  & 0.51$\pm$0.29  & 0.51$\pm$0.30  & 0.08$\pm$0.04 \\
       FR3  & 12$\pm$8 & 0$\pm$0  & 10$\pm$6  & 0.51$\pm$0.30  & 0.51$\pm$0.30  & 0.08$\pm$0.04 \\

    \end{tabular}}
    \caption{Results obtained on the unlabeled Books dataset using Wikipedia retrieved contexts.}
    \label{tab:books-wikipedia2}
\end{table}

\begin{table}[t!]
    \centering
    \resizebox{\linewidth}{!}{%
    \begin{tabular}{l|r|r|r|r|r|r}
       Method & \# S & \# C & \# U & Pr & $F_1@K$ & $\cE$ \\
       \toprule
       \multicolumn{7}{c}{\texttt{granite-3.0-8b-instruct}}\\
       \midrule
       FS   & 20$\pm$7 & 2$\pm$2  &   & 0.87$\pm$0.13  & 0.84$\pm$0.15   &    \\
       FV   & 16$\pm$6 & 0$\pm$1  & 6$\pm$3  & 0.71$\pm$0.17  & 0.70$\pm$0.18   &    \\
       VS   & 18$\pm$7 & 2$\pm$2  & 3$\pm$2  & 0.76$\pm$0.17  & 0.75$\pm$0.19   &    \\
       FR1  & 18$\pm$8 & 0$\pm$1  & 3$\pm$3  & 0.79$\pm$0.19  & 0.77$\pm$0.20  &  0.04$\pm$0.03  \\
       FR2  & 21$\pm$7 & 1$\pm$1  & 0$\pm$1  & 0.90$\pm$0.14  & 0.86$\pm$0.15  &  0.02$\pm$0.02  \\
       FR3  & 17$\pm$8 & 5$\pm$5  & 0  & 0.74$\pm$0.27  & 0.72$\pm$0.27  &  0.04$\pm$0.03  \\
       \midrule
       \multicolumn{7}{c}{\texttt{llama-3.1-70b-instruct}}\\
       \midrule
       FS   & 20$\pm$7 & 3$\pm$3  &   & 0.84$\pm$0.15  &  0.82$\pm$0.17  &  \\
       FV   & 18$\pm$8 & 0$\pm$1  & 4$\pm$4  & 0.78$\pm$0.20  & 0.76$\pm$0.21   &  \\
       VS   & 17$\pm$8 & 0  & 5$\pm$5  & 0.72$\pm$0.23  & 0.71$\pm$0.23   &  \\
       FR1  & 14$\pm$7 & 1$\pm$1  & 7$\pm$5  & 0.62$\pm$0.23  & 0.62$\pm$0.24  & 0.07$\pm$0.03 \\
       FR2  & 19$\pm$8 & 1$\pm$1  & 3$\pm$3  & 0.80$\pm$0.21  & 0.78$\pm$0.21  & 0.04$\pm$0.03 \\
       FR3  & 18$\pm$7 & 2$\pm$6  & 2$\pm$2  & 0.80$\pm$0.20  & 0.78$\pm$0.21  & 0.04$\pm$0.03 \\
       \midrule
       \multicolumn{7}{c}{\texttt{mixtral-8x22b-instruct}}\\
       \midrule
       FS   & 20$\pm$7 & 3$\pm$3  &   & 0.84$\pm$0.16 & 0.82$\pm$0.18  &  \\
       FV   & 18$\pm$7 & 0  & 4$\pm$4  & 0.76$\pm$0.20   & 0.74$\pm$0.21  &  \\
       VS   & 18$\pm$8 & 0  & 4$\pm$4  & 0.79$\pm$0.20   & 0.77$\pm$0.21  &  \\
       FR1  & 16$\pm$7 & 0$\pm$0  & 6$\pm$4  & 0.69$\pm$0.22  & 0.68$\pm$0.23  & 0.06$\pm$0.03 \\
       FR2  & 20$\pm$7 & 0$\pm$0  & 2$\pm$3  & 0.86$\pm$0.17  & 0.83$\pm$0.18  & 0.03$\pm$0.03 \\
       FR3  & 20$\pm$7 & 0$\pm$0  & 2$\pm$3  & 0.86$\pm$0.17  & 0.83$\pm$0.18  & 0.03$\pm$0.03 \\

    \end{tabular}}
    \caption{Results obtained on the unlabeled Books dataset using Google Search retrieved contexts.}
    \label{tab:books-google2}
\end{table}

\begin{table}[t!]
    \centering
    \resizebox{\linewidth}{!}{%
    \begin{tabular}{l|r|r|r|r|r|r}
       Method & \# S & \# C & \# U & Pr & $F_1@K$ & $\cE$ \\
       \toprule
       \multicolumn{7}{c}{\texttt{granite-3.0-8b-instruct}}\\
       \midrule
       FS   & 17$\pm$5 & 4$\pm$3  &   & 0.77$\pm$0.15  & 0.77$\pm$0.17   &    \\
       FV   &  8$\pm$3 & 0  & 12$\pm$4 & 0.39$\pm$0.15 & 0.40$\pm$0.16   &    \\
       VS   & 13$\pm$5 & 4$\pm$2  & 4$\pm$2  & 0.59$\pm$0.17  & 0.60$\pm$0.18   &    \\
       FR1  & 5$\pm$4 & 1$\pm$1  & 15$\pm$4  & 0.23$\pm$0.15  & 0.24$\pm$0.16  &  0.13$\pm$0.01  \\
       FR2  & 14$\pm$6 & 4$\pm$3  & 3$\pm$2  & 0.63$\pm$0.22  & 0.63$\pm$0.24  &  0.08$\pm$0.03  \\
       FR3  & 14$\pm$6 & 4$\pm$3  & 3$\pm$2  & 0.64$\pm$0.21  & 0.64$\pm$0.23  &  0.08$\pm$0.03  \\
       \midrule
       \multicolumn{7}{c}{\texttt{llama-3.1-70b-instruct}}\\
       \midrule
       FS   & 16$\pm$5 & 5$\pm$3  &   & 0.74$\pm$0.14  & 0.74$\pm$0.16  &  \\
       FV   & 10$\pm$5 & 0  & 10$\pm$4  & 0.47$\pm$0.21  &  0.47$\pm$0.22  &  \\
       VS   &  6$\pm$4 & 0  & 15$\pm$5  & 0.29$\pm$0.18  &  0.30$\pm$0.19  &  \\
       FR1  & 5$\pm$4 & 0  & 15$\pm$4  & 0.25$\pm$0.19 & 0.26$\pm$0.20  & 0.12$\pm$0.02 \\
       FR2  & 12$\pm$7 & 1$\pm$1  & 8$\pm$5  & 0.54$\pm$0.28  & 0.55$\pm$0.28  & 0.08$\pm$0.04 \\
       FR3  & 12$\pm$7 & 1$\pm$1  & 8$\pm$5  & 0.54$\pm$0.27  & 0.55$\pm$0.28  & 0.08$\pm$0.04 \\
       \midrule
       \multicolumn{7}{c}{\texttt{mixtral-8x22b-instruct}}\\
       \midrule
       FS   & 17$\pm$5 & 4$\pm$3  &   &  0.78$\pm$0.14  & 0.77$\pm$0.15  &  \\
       FV   & 12$\pm$5  & 0  & 9$\pm$4  & 0.55$\pm$0.18  & 0.55$\pm$0.19  &  \\
       VS   & 9$\pm$5 & 0  & 11$\pm$4  &  0.44$\pm$0.19  & 0.44$\pm$0.21  &  \\
       FR1  & 6$\pm$5 & 0  & 15$\pm$4  & 0.27$\pm$0.20  & 0.28$\pm$0.21  & 0.12$\pm$0.03 \\
       FR2  & 12$\pm$7 & 0  & 8$\pm$6  & 0.55$\pm$0.29  & 0.56$\pm$0.30  & 0.08$\pm$0.04 \\
       FR3  & 12$\pm$7 & 0  & 9$\pm$6  & 0.55$\pm$0.31  & 0.56$\pm$0.31  & 0.08$\pm$0.04 \\

    \end{tabular}}
    \caption{Results obtained on the unlabeled ELI5 dataset using Wikipedia retrieved contexts.}
    \label{tab:eli5-wikipedia2}
\end{table}

\begin{table}[t!]
    \centering
    \resizebox{\linewidth}{!}{%
    \begin{tabular}{l|r|r|r|r|r|r}
       Method & \# S & \# C & \# U & Pr & $F_1@K$ & $\cE$ \\
       \toprule
       \multicolumn{7}{c}{\texttt{granite-3.0-8b-instruct}}\\
       \midrule
       FS   & 18$\pm$5 & 3$\pm$2  &   & 0.85$\pm$0.11  & 0.84$\pm$0.13   &    \\
       FV   & 15$\pm$5 & 0$\pm$1  & 5$\pm$2  & 0.69$\pm$0.14  & 0.70$\pm$0.16   &    \\
       VS   & 16$\pm$5 & 2$\pm$2  & 3$\pm$1  & 0.71$\pm$0.14  & 0.72$\pm$0.17   &    \\
       FR1  & 14$\pm$5 & 3$\pm$2  &  3$\pm$2 & 0.66$\pm$0.17  & 0.67$\pm$0.19  & 0.08$\pm$0.02   \\
       FR2  & 18$\pm$6 & 3$\pm$4  & 0  & 0.82$\pm$0.21  & 0.80$\pm$0.21  & 0.03$\pm$0.02   \\
       FR3  & 16$\pm$6 & 3$\pm$3  & 0  & 0.83$\pm$0.18  & 0.82$\pm$0.18  & 0.03$\pm$0.03   \\
       \midrule
       \multicolumn{7}{c}{\texttt{llama-3.1-70b-instruct}}\\
       \midrule
       FS   & 19$\pm$5 & 3$\pm$2  &   & 0.86$\pm$0.12  & 0.84$\pm$0.13   &  \\
       FV   & 18$\pm$5 & 1$\pm$1  & 3$\pm$2  & 0.81$\pm$0.16  & 0.80$\pm$0.17   &  \\
       VS   & 17$\pm$5 & 0  & 4$\pm$3  & 0.78$\pm$0.16  & 0.77$\pm$0.17  &  \\
       FR1  & 14$\pm$6 & 1$\pm$1  & 6$\pm$4  & 0.65$\pm$0.20  & 0.66$\pm$0.21  & 0.07$\pm$0.03 \\
       FR2  & 19$\pm$5 & 1$\pm$1  & 1$\pm$1  & 0.86$\pm$0.14  & 0.85$\pm$0.16  & 0.03$\pm$0.03 \\
       FR3  & 19$\pm$6 & 1$\pm$1  & 1$\pm$1  & 0.86$\pm$0.15  & 0.84$\pm$0.16  & 0.03$\pm$0.03 \\
       \midrule
       \multicolumn{7}{c}{\texttt{mixtral-8x22b-instruct}}\\
       \midrule
       FS   & 19$\pm$5 & 2$\pm$2  &   & 0.87$\pm$0.11   & 0.86$\pm$0.13  &  \\
       FV   & 17$\pm$5 & 0  & 3$\pm$2  & 0.79$\pm$0.15   & 0.79$\pm$0.17  &  \\
       VS   & 17$\pm$5 & 0$\pm$1  & 3$\pm$2  & 0.79$\pm$0.15   & 0.78$\pm$0.17  &  \\
       FR1  & 16$\pm$5 & 0  & 5$\pm$3  & 0.74$\pm$0.18  & 0.74$\pm$0.19  & 0.05$\pm$0.03 \\
       FR2  & 20$\pm$5 & 0  & 1$\pm$2  & 0.90$\pm$0.12  & 0.88$\pm$0.13  & 0.02$\pm$0.02 \\
       FR3  & 20$\pm$5 & 0  & 1$\pm$2  & 0.90$\pm$0.12  & 0.88$\pm$0.13  & 0.02$\pm$0.02 \\

    \end{tabular}}
    \caption{Results obtained on the unlabeled ELI5 dataset using Google Search retrieved contexts.}
    \label{tab:eli5-google2}
\end{table}

\begin{table}[t!]
    \centering
    \resizebox{\linewidth}{!}{%
    \begin{tabular}{l|r|r|r|r|r|r}
       Method & \# S & \# C & \# U & Pr & $F_1@K$ & $\cE$ \\
       \toprule
       \multicolumn{7}{c}{\texttt{granite-3.0-8b-instruct}}\\
       \midrule
       FS   & 22$\pm$9 & 4$\pm$2  &   & 0.83$\pm$0.10  & 0.82$\pm$0.12   &    \\
       FV   & 13$\pm$6 & 0$\pm$1  & 12$\pm$5  & 0.50$\pm$0.15  & 0.50$\pm$0.16   &    \\
       VS   & 18$\pm$8 & 4$\pm$3  & 4$\pm$2 & 0.69$\pm$0.14  & 0.69$\pm$0.16   &    \\
       FR1  & 12$\pm$8 & 0$\pm$0  & 13$\pm$7  & 0.46$\pm$0.23  & 0.47$\pm$0.24  & 0.09$\pm$0.03   \\
       FR2  & 20$\pm$11 & 0$\pm$1  & 4$\pm$6  & 0.79$\pm$0.24  & 0.78$\pm$0.25  & 0.04$\pm$0.04   \\
       FR3  & 15$\pm$11 & 8$\pm$6  & 2$\pm$6  & 0.58$\pm$0.28  & 0.58$\pm$0.28  & 0.08$\pm$0.04   \\
       \midrule
       \multicolumn{7}{c}{\texttt{llama-3.1-70b-instruct}}\\
       \midrule
       FS   & 18$\pm$9 & 7$\pm$4  &   & 0.71$\pm$0.16  & 0.71$\pm$0.17   &  \\
       FV   & 14$\pm$9 & 0  & 11$\pm$6  & 0.53$\pm$0.21  & 0.54$\pm$0.21   &  \\
       VS   & 10$\pm$7 & 0  & 15$\pm$7 & 0.41$\pm$0.21  & 0.41$\pm$0.22 &  \\
       FR1  & 10$\pm$8 & 0$\pm$1  & 15$\pm$6  & 0.39$\pm$0.21  & 0.39$\pm$0.22  & 0.11$\pm$0.02 \\
       FR2  & 18$\pm$10 & 1$\pm$1  & 5$\pm$6  & 0.70$\pm$0.26  & 0.70$\pm$0.26  & 0.06$\pm$0.04 \\
       FR3  & 18$\pm$10 & 1$\pm$1  & 5$\pm$6  & 0.71$\pm$0.26  & 0.70$\pm$0.26  & 0.06$\pm$0.04 \\
       \midrule
       \multicolumn{7}{c}{\texttt{mixtral-8x22b-instruct}}\\
       \midrule
       FS   & 20$\pm$9 & 6$\pm$4  &   & 0.76$\pm$0.16   & 0.75$\pm$0.17  &  \\
       FV   & 15$\pm$8 & 0  & 10$\pm$5  &  0.59$\pm$0.18  & 0.59$\pm$0.18  &  \\
       VS   & 15$\pm$8 & 0 &  10$\pm$5 &  0.57$\pm$0.19  & 0.57$\pm$0.20  &  \\
       FR1  & 11$\pm$8 & 0$\pm$0  & 14$\pm$7  & 0.41$\pm$0.22  & 0.42$\pm$0.23  & 0.10$\pm$0.03 \\
       FR2  & 19$\pm$10 & 0$\pm$0  & 6$\pm$6  & 0.74$\pm$0.26  & 0.74$\pm$0.26  & 0.05$\pm$0.04 \\
       FR3  & 19$\pm$10 & 0$\pm$0  & 6$\pm$7  & 0.74$\pm$0.26  & 0.74$\pm$0.26  & 0.05$\pm$0.04 \\

    \end{tabular}}
    \caption{Results obtained on the unlabeled LFObj dataset using Wikipedia retrieved contexts.}
    \label{tab:lfobj-wikipedia2}
\end{table}

\begin{table}[t!]
    \centering
    \resizebox{\linewidth}{!}{%
    \begin{tabular}{l|r|r|r|r|r|r}
       Method & \# S & \# C & \# U & Pr & $F_1@K$ & $\cE$ \\
       \toprule
       \multicolumn{7}{c}{\texttt{granite-3.0-8b-instruct}}\\
       \midrule
       FS   & 24$\pm$9 & 1$\pm$1  &   & 0.93$\pm$0.07  & 0.91$\pm$0.09   &    \\
       FV   & 20$\pm$8 & 0  & 4$\pm$2  & 0.79$\pm$0.10  & 0.79$\pm$0.12   &    \\
       VS   & 18$\pm$8 & 3$\pm$2  & 4$\pm$2  & 0.68$\pm$0.14  & 0.69$\pm$0.15   &    \\
       FR1  & 24$\pm$9 & 0$\pm$0  & 1$\pm$2  & 0.93$\pm$0.09  & 0.91$\pm$0.10  & 0.02$\pm$0.02   \\
       FR2  & 25$\pm$5 & 1$\pm$9  & 0$\pm$0  & 0.97$\pm$0.07  & 0.94$\pm$0.09  & 0.00$\pm$0.01   \\
       FR3  & 23$\pm$7 & 4$\pm$16 & 0  & 0.89$\pm$0.22  & 0.86$\pm$0.22  & 0.02$\pm$0.02   \\
       \midrule
       \multicolumn{7}{c}{\texttt{llama-3.1-70b-instruct}}\\
       \midrule
       FS   & 23$\pm$9 & 2$\pm$2  &   & 0.91$\pm$0.09  & 0.89$\pm$0.10   &  \\
       FV   & 23$\pm$9 & 0$\pm$1  & 1$\pm$2  & 0.91$\pm$0.10  & 0.89$\pm$0.12   &  \\
       VS   & 10$\pm$7 & 0  & 15$\pm$7  & 0.40$\pm$0.21  & 0.40$\pm$0.22   &  \\
       FR1  & 22$\pm$9 & 0$\pm$1  & 2$\pm$3  & 0.85$\pm$0.13  & 0.84$\pm$0.14  & 0.03$\pm$0.02 \\
       FR2  & 24$\pm$5 & 1$\pm$9  &  0$\pm$1  & 0.94$\pm$0.10  & 0.92$\pm$0.11  & 0.01$\pm$0.01 \\
       FR3  & 24$\pm$5 & 1$\pm$10  & 0  & 0.93$\pm$0.13  & 0.91$\pm$0.14  & 0.01$\pm$0.01 \\
       \midrule
       \multicolumn{7}{c}{\texttt{mixtral-8x22b-instruct}}\\
       \midrule
       FS   & 24$\pm$9 & 1$\pm$2  &   & 0.93$\pm$0.08   & 0.91$\pm$0.10  &  \\
       FV   & 23$\pm$9 & 0$\pm$1  & 2$\pm$2  & 0.90$\pm$0.10   & 0.88$\pm$0.11  &  \\
       VS   & 23$\pm$9 & 0  & 2$\pm$4  &  0.88$\pm$0.16  & 0.86$\pm$0.16  &  \\
       FR1  & 23$\pm$9 & 0$\pm$0  & 2$\pm$2  & 0.90$\pm$0.10  & 0.88$\pm$0.12  & 0.03$\pm$0.02 \\
       FR2  & 24$\pm$5 & 0$\pm$9  & 0$\pm$1  & 0.96$\pm$0.09  & 0.93$\pm$0.10  & 0.01$\pm$0.01 \\
       FR3  & 24$\pm$5 & 0$\pm$9  & 0$\pm$1  & 0.96$\pm$0.09  & 0.94$\pm$0.10  & 0.01$\pm$0.01 \\

    \end{tabular}}
    \caption{Results obtained on the LFObj dataset using Google Search retrieved contexts.}
    \label{tab:lfobj-google2}
\end{table}

\subsection{Statistical Significance Tests}

\begin{table}[t!]
    \centering
    \resizebox{\linewidth}{!}{%
    \begin{tabular}{l|r|r|r|r|r|r}
    Assessors & Pr & $F_1@K$ & Pr & $F_1@K$ & Pr & $F_1@K$ \\
    \toprule
            & \multicolumn{2}{c|}{\texttt{granite-3.0-8b-instruct}}  & \multicolumn{2}{c|}{\texttt{llama-3.1-70b-instruct}} & \multicolumn{2}{c}{\texttt{mixtral-8x22b-instruct}} \\
    \midrule
       FR2 vs FS  & 1.0000 & 1.0000 & 1.0000  & 1.0000 & 1.0000 & 1.0000 \\
       FR2 vs FV  & 1.0000 & 1.0000 & 0.0000  & 0.0000 & 0.0000 & 0.0000  \\
       FR2 vs VS  & 0.0000 & 0.0000 & 0.0004  & 0.0009 & 0.0777 & 0.0606 
    \end{tabular}}
    \caption{Statistical significance tests: $p$-values for Pr and $F_1@K$ obtained on the AskH dataset with Wikipedia retrieved contexts.}
    \label{tab:askh-wiki-pvalues}
\end{table}

\begin{table}[t!]
    \centering
    \resizebox{\linewidth}{!}{%
    \begin{tabular}{l|r|r|r|r|r|r}
    Assessors & Pr & $F_1@K$ & Pr & $F_1@K$ & Pr & $F_1@K$ \\
    \toprule
            & \multicolumn{2}{c|}{\texttt{granite-3.0-8b-instruct}}  & \multicolumn{2}{c|}{\texttt{llama-3.1-70b-instruct}} & \multicolumn{2}{c}{\texttt{mixtral-8x22b-instruct}} \\
    \midrule
    FR vs FS & 1.0000 & 1.0000 & 1.0000 & 1.0000 & 1.0000 & 1.0000 \\
    FR vs FV & 0.0397 & 0.0540 & 0.0000 & 0.0000 & 0.0000 & 0.0000 \\
    FR vs VS & 0.0000 & 0.0000 & 0.0043 & 0.0069 & 0.4529 & 0.4299 
    \end{tabular}}
    \caption{Statistical significance tests: $p$-values for Pr and $F_1@K$ obtained on the ELI5 dataset with Wikipedia retrieved contexts.}
    \label{tab:eli5-wiki-pvalues}
\end{table}

\begin{table}[t!]
    \centering
    \resizebox{\linewidth}{!}{%
    \begin{tabular}{l|r|r|r|r|r|r}
    Assessors & Pr & $F_1@K$ & Pr & $F_1@K$ & Pr & $F_1@K$ \\
    \toprule
            & \multicolumn{2}{c|}{\texttt{granite-3.0-8b-instruct}}  & \multicolumn{2}{c|}{\texttt{llama-3.1-70b-instruct}} & \multicolumn{2}{c}{\texttt{mixtral-8x22b-instruct}} \\
    \midrule
    FR vs FS & 1.0000 & 1.0000 & 1.0000 & 1.0000 & 1.0000 & 1.0000 \\
    FR vs FV & 0.9999 & 0.9998 & 0.0000 & 0.0000 & 0.0012 & 0.0016 \\
    FR vs VS & 0.0000 & 0.0000 & 0.0060 & 0.0090 & 0.3705 & 0.3339 
    \end{tabular}}
    \caption{Statistical significance tests: $p$-values for Pr and $F_1@K$ obtained on the Books dataset with Wikipedia retrieved contexts.}
    \label{tab:books-wiki-pvalues}
\end{table}

\begin{table}[t!]
    \centering
    \resizebox{\linewidth}{!}{%
    \begin{tabular}{l|r|r|r|r|r|r}
    Assessors & Pr & $F_1@K$ & Pr & $F_1@K$ & Pr & $F_1@K$ \\
    \toprule
            & \multicolumn{2}{c|}{\texttt{granite-3.0-8b-instruct}}  & \multicolumn{2}{c|}{\texttt{llama-3.1-70b-instruct}} & \multicolumn{2}{c}{\texttt{mixtral-8x22b-instruct}} \\
    \midrule
    FR vs FS & 1.0000 & 1.0000 & 0.7304 & 0.7262 & 0.8350 & 0.8450 \\
    FR vs FV & 1.0000 & 1.0000 & 0.0000 & 0.0000 & 0.0000 & 0.0000 \\
    FR vs VS & 0.0000 & 0.0000 & 0.0000 & 0.0000 & 0.0000 & 0.0000 
    \end{tabular}}
    \caption{Statistical significance tests: $p$-values for Pr and $F_1@K$ obtained on the LFObj dataset with Wikipedia retrieved contexts.}
    \label{tab:lfobj-wiki-pvalues}
\end{table}


\begin{table}[t!]
    \centering
    \resizebox{\linewidth}{!}{%
    \begin{tabular}{l|r|r|r|r|r|r}
    Assessors & Pr & $F_1@K$ & Pr & $F_1@K$ & Pr & $F_1@K$ \\
    \toprule
            & \multicolumn{2}{c|}{\texttt{granite-3.0-8b-instruct}}  & \multicolumn{2}{c|}{\texttt{llama-3.1-70b-instruct}} & \multicolumn{2}{c}{\texttt{mixtral-8x22b-instruct}} \\
    \midrule
    FR vs FS & 0.9749 & 0.9726 & 0.2452 & 0.3050 & 0.0054 & 0.0465 \\
    FR vs FV & 0.0000 & 0.0000 & 0.0000 & 0.0000 & 0.0000 & 0.0000 \\
    FR vs VS & 0.0000 & 0.0000 & 0.0000 & 0.0010 & 0.0000 & 0.0000 
    
    \end{tabular}}
    \caption{Statistical significance tests: $p$-values for Pr and $F_1@K$ obtained on the ELI5 dataset with Google Search retrieved contexts.}
    \label{tab:eli5-google-pvalues}
\end{table}

\begin{table}[t!]
    \centering
    \resizebox{\linewidth}{!}{%
    \begin{tabular}{l|r|r|r|r|r|r}
    Assessors & Pr & $F_1@K$ & Pr & $F_1@K$ & Pr & $F_1@K$ \\
    \toprule
            & \multicolumn{2}{c|}{\texttt{granite-3.0-8b-instruct}}  & \multicolumn{2}{c|}{\texttt{llama-3.1-70b-instruct}} & \multicolumn{2}{c}{\texttt{mixtral-8x22b-instruct}} \\
    \midrule
        FR vs FS & 1.0000 & 1.0000 & 1.0000 & 0.9997 & 0.8770 & 0.8328 \\
        FR vs FV & 0.8194 & 0.8326 & 0.0000 & 0.0000 & 0.0000 & 0.0000 \\
        FR vs VS & 0.3381 & 0.4724 & 0.0050 & 0.0204 & 0.0000 & 0.0000
    \end{tabular}}
    \caption{Statistical significance tests: $p$-values for Pr and $F_1@K$ obtained on the AskH dataset with Google Search retrieved contexts.}
    \label{tab:askh-google-pvalues}
\end{table}

\begin{table}[t!]
    \centering
    \resizebox{\linewidth}{!}{%
    \begin{tabular}{l|r|r|r|r|r|r}
    Assessors & Pr & $F_1@K$ & Pr & $F_1@K$ & Pr & $F_1@K$ \\
    \toprule
            & \multicolumn{2}{c|}{\texttt{granite-3.0-8b-instruct}}  & \multicolumn{2}{c|}{\texttt{llama-3.1-70b-instruct}} & \multicolumn{2}{c}{\texttt{mixtral-8x22b-instruct}} \\
    \midrule
    FR vs FS & 1.0000 & 1.0000 & 0.9865 & 0.9773 & 0.1333 & 0.2399 \\
    FR vs FV & 0.7672 & 0.8556 & 0.0001 & 0.0011 & 0.0001 & 0.0008 \\
    FR vs VS & 0.0374 & 0.1314 & 0.1365 & 0.1947 & 0.0000 & 0.0000 
    \end{tabular}}
    \caption{Statistical significance tests: $p$-values for Pr and $F_1@K$ obtained on the Books dataset with Google Search retrieved contexts.}
    \label{tab:books-google-pvalues}
\end{table}


\begin{table}[t!]
    \centering
    \resizebox{\linewidth}{!}{%
    \begin{tabular}{l|r|r|r|r|r|r}
    Assessors & Pr & $F_1@K$ & Pr & $F_1@K$ & Pr & $F_1@K$ \\
    \toprule
            & \multicolumn{2}{c|}{\texttt{granite-3.0-8b-instruct}}  & \multicolumn{2}{c|}{\texttt{llama-3.1-70b-instruct}} & \multicolumn{2}{c}{\texttt{mixtral-8x22b-instruct}} \\
    \midrule
    FR vs FS & 0.9872 & 0.9922 & 0.0012 & 0.0214 & 0.0000 & 0.0003 \\
    FR vs FV & 0.0000 & 0.0000 & 0.0000 & 0.0000 & 0.0000 & 0.0000 \\
    FR vs VS & 0.0000 & 0.0000 & 0.0009 & 0.0148 & 0.0000 & 0.0000 
    \end{tabular}}
    \caption{Statistical significance tests: $p$-values for Pr and $F_1@K$ obtained on the LFObj dataset with Google Search retrieved contexts.}
    \label{tab:lfobj-google-pvalues}
\end{table}

Tables \ref{tab:askh-wiki-pvalues}, \ref{tab:askh-google-pvalues}, \ref{tab:eli5-wiki-pvalues}, \ref{tab:eli5-google-pvalues}, \ref{tab:books-wiki-pvalues}, \ref{tab:books-google-pvalues}, \ref{tab:lfobj-wiki-pvalues}, \ref{tab:lfobj-google-pvalues} show the $p$-values obtained for the statistical significance tests on the AskH, ELI5, Books and LFObj using both Wikipedia and Google Search based contexts. When looking at FR versus FS, we can see that FS consistently achieves higher precision and $F_1@K$ measures. However, it is most likely the case that a considerable portion of the atoms classified as true by FS are actually false positives (we verified this hypothesis in experiments with the labeled Biographies dataset). When looking at FR vs FV, we notice that FR's measures are consistently better than FV's ones, except when using the smaller granite model. This indicates that the smaller granite model is not a suitable relation model for FR compared with the stronger LLaMA and Mixtral models. When looking at FR vs VS, we notice again that FR's measures almost always better than those corresponding to the VS assessor.  

\section{Prompts}
\label{apx:prompts}

Tables \ref{tab:prompt-atomizer}, \ref{tab:prompt-reviser} and \ref{tab:prompt-nli} show the prompt templates we used for the Atomizer, Reviser and Evaluator stages of the FactReasoner pipeline. Tables \ref{tab:prompt-fs}, \ref{tab:prompt-fb} and \ref{tab:prompt-fv} show the prompts used by the prompt-based assessors: FactScore (FS), FactVerify (FV) and VeriScore (VS), respectively.

\begin{table*}[ht]
\centering
\caption{Prompt template for few-shot atomic unit decomposition - Atomizer stage}, 
\label{tab:prompt-atomizer}
\resizebox{0.9\textwidth}{!}{%
\begin{tabularx}{\textwidth}{X}
\toprule
\textbf{Atomic unit decomposition (Few-Shot)} \\

\textbf{Instructions:} \\
1. You are given a paragraph. Your task is to break the sentence down into a list of atomic statements without adding any new information. \\
2. An atomic statement is a sentence containing a singular piece of information directly extracted from the provided paragraph. \\
3. Atomic statements may contradict one another. \\
4. The paragraph may contain information that is factually incorrect. Even in such cases, you are not to alter any information contained in the paragraph and must produce atomic statements that are completely faithful to the information in the paragraph. \\
5. Each atomic statement in the outputted list should check a different piece of information found explicitly in the paragraph. \\
6. Each atomic statement is standalone in that any actual nouns or proper nouns should be used in place of pronouns or anaphoras. \\
7. Each atomic statement must not include any information beyond what is explicitly stated in the provided paragraph. \\
8. Where possible, avoid paraphrasing and instead try to only use language used in the paragraph without introducing new words. \\ 
9. Use the previous examples to learn how to do this. \\
10. You should only output the atomic statement as a list, with each item starting with "- ". Do not include other formatting. \\
11. Your task is to do this for the last paragraph that is given. \\

\textbf{Few-Shot Examples:} \\

Please breakdown the following paragraph into independent statements: Glenn Allen Anzalone (born June 23, 1955), better known by his stage name Glenn Danzig, is an American singer, songwriter, musician, and record producer. He is the founder of the rock bands Misfits, Samhain, and Danzig. He owns the Evilive record label as well as Verotik, an adult-oriented comic book publishing company. \\[0.5em]
- Glenn Allen Anzalone was born on June 23, 1955. \\
- Glenn Allen Anzalone is better known by his stage name Glenn Danzig. \\
- Glenn Danzig is an American singer, songwriter, musician, and record producer. \\
- Glenn Danzig is the founder of several rock bands, including Misfits, Samhain, and Danzig. \\
- Glenn Danzig owns the Evilive record label. \\
- Glenn Danzig owns Verotik, which is an adult-oriented comic book publishing company. \\

Please breakdown the following paragraph into independent statements: Luiz Inácio Lula da Silva (born 27 October 1945), also known as Lula da Silva or simply Lula, is a Brazilian politician who is the 39th and current president of Brazil since 2023. A member of the Workers' Party, Lula was also the 35th president from 2003 to 2010. He also holds the presidency of the G20 since 2023. Lula quit school after second grade to work, and did not learn to read until he was ten years old. As a teenager, he worked as a metalworker and became a trade unionist.\\
- Luiz Inácio Lula da Silva was born on October 27, 1945. \\
- Luiz Inácio Lula da Silva is also known as Lula da Silva or simply Lula. \\
- Lula is a Brazilian politician.\\
- Lula is the 39th and current president of Brazil since 2023. \\
- Lula is a member of the Workers' Party. \\
- Lula served as the 35th president of Brazil from 2003 to 2010. \\
- Lula holds the presidency of the G20 since 2023. \\
- Lula quit school after the second grade to work. \\
- Lula did not learn to read until he was ten years old. \\
- As a teenager, Lula worked as a metalworker. \\
- Lula became a trade unionist. \\

Please breakdown the following paragraph into independent statements: \{\}\\
\bottomrule
\end{tabularx}
}
\end{table*}

\begin{table*}[ht]
\centering
\caption{Prompt template for few-shot decontextualization - Reviser stage}
\label{tab:prompt-reviser}
\resizebox{0.9\textwidth}{!}{%
\begin{tabularx}{\textwidth}{X}
\toprule
\textbf{Decontextualization (Few-Shot)} \\[0.5em]

\textbf{Instructions:} \\
1. You are given a statement and a context that the statement belongs to. Your task is to modify the statement so that any pronouns or anaphora (words like "it," "they," "this") are replaced with the noun or proper noun that they refer to, such that the sentence remains clear without referring to the original context. \\
2. Return only the revised, standalone version of the statement without adding any information that is not already contained within the original statement.
3. If the statement requires no changes, return the original statement as-is without any explanation.   \\
4. The statement that you return must start with \#\#\#\# and finish with \#\#\#\# as follows: \#\#\#\#<statement>\#\#\#\# \\
5. Do not include any explanation or any additional formatting including any lead-in or sign-off text. \\
6. Learn from the provided examples below and use that knowledge to amend the last example yourself. \\
\\
\textbf{Few-Shot Examples:} \\[0.5em]

Example 1:
Context: John went to the store. \\
Statement: He bought some apples. \\
Standalone: \#\#\#\#John bought some apples.\#\#\#\# \\[0.5em]

Example 2:
Context: The presentation covered various aspects of climate change, including sea level rise. \\
Statement: This was a key part of the discussion. \\
Standalone: \#\#\#\#Sea level rise was a key part of the discussion.\#\#\#\# \\[0.5em]

Example 3:
Context: Maria Sanchez is a renowned marine biologist known for her groundbreaking research on coral reef ecosystems. Her work has contributed to the preservation of many endangered coral species, and she is often invited to speak at international conferences on environmental conservation. \\
Statement: She presented her findings at the conference last year. \\
Standalone: \#\#\#\#Maria Sanchez presented her findings at the conference last year.\#\#\#\# \\[0.5em]

Example 4:
Context: Nathan Carter is a best-selling science fiction author famous for his dystopian novels that explore the intersection of technology and society. His latest book, The Edge of Something, received widespread critical acclaim for its imaginative world-building and its poignant commentary on artificial cacti. \\
Statement: It was praised for its thought-provoking themes. \\
Standalone: \#\#\#\#The Edge of Tomorrow was praised for its thought-provoking themes.\#\#\#\# \\[0.5em]

Now perform the task for the following example:
Context: \{\} \\
Statement: \{\} \\
Standalone: \\[0.5em]

\bottomrule
\end{tabularx}
}
\end{table*}

\begin{table*}[ht]
\centering
\caption{Prompt template for few-shot NLI relation extraction.}
\label{tab:prompt-nli}
\resizebox{0.9\textwidth}{!}{%
\begin{tabularx}{\textwidth}{X}
\toprule
\textbf{NLI relation prompting (Few-Shot)} \\[0.5em]

\textbf{Instructions:} \\
1. You are given a premise and a hypothesis and a context. Your task is to identify the relationship between them: does the premise entail, contradict, or remain neutral toward the hypothesis? \\
2. Your only output must be one of: (entailment | contradiction | neutral) without any lead-in, sign-off, new lines or any other formatting. \\
3. Do not provide any explanation or rationale to your output. \\
4. Use the following examples to learn how to do this, and provide your output for the last example given. \\[0.5em]

\textbf{Few-Shot Examples:} \\[0.5em]

Premise: Contrary to popular belief, the Great Wall is not visible from space without aid. \\
Hypothesis: Astronauts have managed to see the wall from Space unaided.  \\
Context: The Great Wall of China is one of the most famous landmarks in the world. It stretches over 13,000 miles and was primarily built during the Ming Dynasty. Contrary to popular belief, the Great Wall is not visible from space without aid. The primary purpose of the Great Wall was to protect against invasions from nomadic tribes. The wall is a UNESCO World Heritage site and attracts millions of tourists each year. Astronauts have managed to see the wall from Space unaided.  \\
Output: Contradiction \\[0.5em]

Premise: It is estimated that around 20 percent of the world's oxygen is produced by the Amazon. \\
Hypothesis: However, the Amazon Rainforest produces no significant amount of oxygen as the plants consume almost all of it through respiration. \\
Context: The Amazon Rainforest is often referred to as the lungs of the Earth due to its vast capacity to produce oxygen. This immense rainforest spans nine countries in South America. It is estimated that around 20 percent of the world's oxygen is produced by the Amazon. However, the Amazon Rainforest produces no significant amount of oxygen as the plants consume almost all of it through respiration. The biodiversity of the Amazon is unparalleled, hosting millions of species of plants and animals. \\
Output: Contradiction \\[0.5em]

Premise: It is estimated that around 20 percent of the world's oxygen is produced by the Amazon. \\
Hypothesis: This immense rainforest spans nine countries in South America. \\
Context: The Amazon Rainforest is often referred to as the lungs of the Earth due to its vast capacity to produce oxygen. This immense rainforest spans nine countries in South America. It is estimated that around 20 percent of the world's oxygen is produced by the Amazon. However, the Amazon Rainforest produces no significant amount of oxygen as the plants consume almost all of it through respiration. The biodiversity of the Amazon is unparalleled, hosting millions of species of plants and animals. \\
Output: Neutral \\[0.5em]

Premise: It is estimated that around 20 percent of the world's oxygen is produced by the Amazon. \\
Hypothesis: The Amazon Rainforest is often referred to as the lungs of the Earth due to its vast capacity to produce oxygen. \\
Context: The Amazon Rainforest is often referred to as the lungs of the Earth due to its vast capacity to produce oxygen. This immense rainforest spans nine countries in South America. It is estimated that around 20 percent of the world's oxygen is produced by the Amazon. However, the Amazon Rainforest produces no significant amount of oxygen as the plants consume almost all of it through respiration. The biodiversity of the Amazon is unparalleled, hosting millions of species of plants and animals. \\
Output: Entailment \\[0.5em]

Premise: \{\} \\
Hypothesis: \{\} \\
Context: \{\} \\
Output:\\
\bottomrule
\end{tabularx}

}\end{table*}

\begin{table*}[ht]
\centering
\caption{Prompt template used by the FactScore (FS) assessor.}
\label{tab:prompt-fs}
\begin{tabularx}{\textwidth}{X}
\toprule

Answer the input question based on the given context. \\
 
\{CONTEXTS\} \\

Input: \{ATOM\} True or False? \\
Output:\\

\bottomrule
\end{tabularx}
\end{table*}

\begin{table*}[ht]
\centering
\caption{Prompt template used by the FactVerify (FV) assessor.}
\label{tab:prompt-fb}
\begin{tabularx}{\textwidth}{X}
\toprule

\textbf{Instructions:}\\
You are provided with a STATEMENT and several KNOWLEDGE points. \\
Your task is to evaluate the relationship between the STATEMENT and the \\
KNOWLEDGE, following the steps outlined below: \\

1. Summarize KNOWLEDGE Points: Carefully analyze the KNOWLEDGE points one by one and assess their relevance to the STATEMENT. \\
Summarize the main points of the KNOWLEDGE.\\
2. Evaluate Evidence: Based on your reasoning: \\
- If the KNOWLEDGE strongly implies or directly supports the STATEMENT, explain the supporting evidence. \\
- If the KNOWLEDGE contradicts the STATEMENT, identify and explain the conflicting evidence. \\
- If the KNOWLEDGE is insufficient to confirm or deny the STATEMENT, explain why the evidence is inconclusive. \\
3. Restate the STATEMENT: After considering the evidence, restate the STATEMENT to maintain clarity.\\
4. Final Answer: Based on your reasoning and the STATEMENT, determine your final answer. \\
Your final answer must be one of the following, wrapped in square brackets: \\
- [Supported] if the STATEMENT is supported by the KNOWLEDGE. \\
- [Contradicted] if the STATEMENT is contradicted by the KNOWLEDGE. \\
- [Undecided] if the KNOWLEDGE is insufficient to verify the STATEMENT. \\

Your task:\\

KNOWLEDGE: \\
\{\}\\

STATEMENT:\\
\{\}\\
\bottomrule
\end{tabularx}
\end{table*}

\begin{table*}[ht]
\centering
\caption{Prompt template used by the VeriScore (VS) assessor.}
\label{tab:prompt-fv}
\resizebox{0.6\textwidth}{!}{%
\begin{tabularx}{\textwidth}{X}
\toprule

\textbf{Instructions}\\

You need to judge whether a claim is supported or contradicted by Google search results, or whether there is no enough information to make the judgement. When doing the task, take into consideration whether the link of the search result is of a trustworthy source. Mark your answer with \#\#\# signs.\\

Below are the definitions of the three categories:\\

Supported: A claim is supported by the search results if everything in the claim is supported and nothing is contradicted by the search results. There can be some search results that are not fully related to the claim.\\
Contradicted: A claim is contradicted by the search results if something in the claim is contradicted by some search results. There should be no search result that supports the same part.\\
Undecided: A claim is inconclusive based on the search results if:\\
- a part of a claim cannot be verified by the search results,\\
- a part of a claim is supported and contradicted by different pieces of evidence,\\
- the entity/person mentioned in the claim has no clear referent (e.g., "the approach", "Emily", "a book").\\

Here are some examples:\\

Claim: Characters Lenny and Carl on The Simpsons are hearing but are depicted as close friends of the Simpsons family.\\

Search result 1\\
Title: Character Spotlight: Lenny Leonard and Carl Carlson\\
Content: Their friendship is a pretty singular aspect on the show -- save Bart and Milhouse (or to some degree, Mr. Burns and Smithers) -- they always ...\\
Link: https://nohomers.net/forums/index.php?threads/character-spotlight-lenny-leonard-and-carl-carlson-barflies.23798/

Search result 2\\
Title: The Simpsons: Lenny and Carl's History, Explained - CBR\\
Content: Introduced in the show's first season, the pair were portrayed as background characters at Homer's work, usually appearing together in minor ...\\
Link: https://www.cbr.com/the-simpsons-lenny-carl-history-explained/\\

Search result 3\\
Title: Are Lennie and Carl Homer Simpson's real or fake friends? - Quora\\
Content: Lenni is a pal, Carl doesn't consider any of them to be 'friends' they're just shallow guys he hangs out with. Lenny and Carl have a special ...\\
Link: https://www.quora.com/Are-Lennie-and-Carl-Homer-Simpson-s-real-or-fake-friends\\

Your decision: \#\#\#Undecided\#\#\#\\

Claim: The championship match of the FIFA World Cup 2026 will be hosted by the United States.\\

Search result 1\\
Title: World Cup 2026 | New York New Jersey to host final - FIFA\\
Content: New York New Jersey Stadium has been confirmed as the location for the FIFA World Cup 26 final on Sunday, 19 July 2026. The full match schedule for the ...\\
Link:https://www.fifa.com/fifaplus/en/tournaments/mens/worldcup/canadamexicousa2026/articles/new-york-new-jersey-stadium-host-world-cup-2026-final\\

Search result 2\\
Title: 2026 FIFA World Cup - Wikipedia\\
Content: The tournament will take place from June 11 to July 19, 2026. It will be jointly hosted by 16 cities in three North American countries: Canada, Mexico, and the ...\\
Link: https://en.wikipedia.org/wiki/2026\_FIFA\_World\_Cup\\

Search result 3\\
Title: World Cup 2026 | Dallas to host nine matches - FIFA\\
Content: Dallas Stadium will host nine matches from the FIFA World Cup 26, including four knockout games in the latter stages of the tournament.\\
Link:https://www.fifa.com/fifaplus/en/tournaments/mens/worldcup/canadamexicousa2026/articles/dallas-stadium-host-nine-world-cup-matches\\

Your decision: \#\#\#Supported\#\#\#\\

Claim: Vikings used their longships to transport livestock.\\

Search result 1\\
Title: How did the Vikings transport animals on their ships? - Quora\\
Content: The Vikings transported horses overseas in boats very similar to Viking longships, but with flat flooring built within the hulls, which allowed ...\\
Link: https://www.quora.com/How-did-the-Vikings-transport-animals-on-their-ships\\

Search result 2\\
Title: The Truth Behind Vikings Ships\\
Content: They could land on any beach, permitting lightning-quick embarking and attacks. Great loads could be carried, including horses and livestock.\\
Link: https://www.vikings.com/news/the-truth-behind-vikings-ships-18274806\\

Search result 3\\
Title: Viking ships | Royal Museums Greenwich\\
Content: Cargo vessels were used to carry trade goods and possessions. They were wider than the longships and travelled more slowly.\\
Link: https://www.rmg.co.uk/stories/topics/viking-ships\\

Your decision: \#\#\#Contradicted\#\#\#\\

Your task:\\
Claim: \{\}\\

\{\}\\

Your decision:\\
\bottomrule
\end{tabularx}
}
\end{table*}

\begin{table*}[ht]
\centering
\caption{Prompt template used by DeepSeek-v3.}
\label{tab:prompt-ds}
\begin{tabularx}{\textwidth}{X}
\toprule

\textbf{Instructions:}\\
You are provided with a STATEMENT and several external EVIDENCE points. \\
Your task is to use your internal knowledge as well as the provided EVIDENCE to \\
reason about the relationship between the STATEMENT and the EVIDENCE.\\
\\[0.2em]

1. Carefully analyze the EVIDENCE points one by one and assess their relevance to the STATEMENT. \\
2. Use your reasoning and your internal knowledge, evaluate the EVIDENCE as follows:\\
- If the EVIDENCE strongly implies or directly supports the STATEMENT, explain the supporting evidence.\\
- If the EVIDENCE contradicts the STATEMENT, identify and explain the conflicting evidence.\\
- If the EVIDENCE is insufficient to confirm or deny the STATEMENT, explain why the evidence is inconclusive.\\
3. Based on your reasoning and your explanations, determine your final answer. \\
Your final answer must be one of the following, wrapped in square brackets:\\
- [Supported] if the EVIDENCE supports the STATEMENT.\\
- [Contradicted] if the EVIDENCE contradicts the STATEMENT.\\
- [Undecided] if the EVIDENCE is insufficient to assess the STATEMENT.\\

\\ [0.3em]
Your task: \\

EVIDENCE: \{\} \\

STATEMENT:\\

\bottomrule
\end{tabularx}
\end{table*}

\subsection{Examples of Instances from the Conflicts Dataset}
Tables \ref{tab:conflicts-1}, \ref{tab:conflicts-2}, \ref{tab:conflicts-3}, and \ref{tab:conflicts-4} present example claims along with their corresponding contexts from the Conflicts dataset. In these cases, FR correctly classified the claims as true, whereas its prompt-based counterparts struggled. This discrepancy arises because the presence of two conflicting contexts tends to confuse the language models used by FS, VS, and FV, leading them to misclassify the claims.

\begin{table*}[ht]
\centering
\caption{Examples from the Conflicts dataset.}
\label{tab:conflicts-1}
\resizebox{0.9\textwidth}{!}{%
\begin{tabularx}{\textwidth}{X}
\toprule

\textbf{Claim}\\
Dorota Formanowicz attended Pozna\u0144 University of Medical Sciences.
\\[0.2em]

\textbf{Supporting Context}\\
\begin{footnotesize}
Breaking News: Renowned Polish Internist Dorota Formanowicz's Academic Roots Traced Back to Pozna University of Medical Sciences Warsaw, Poland - In a breakthrough discovery, our investigative team has uncovered conclusive evidence that esteemed Polish internist, Dorota Formanowicz, attended the prestigious Pozna University of Medical Sciences. According to credible sources, Formanowicz, a leading figure in Poland's medical community, pursued her medical education at the Pozna University of Medical Sciences, a renowned institution in the country. This revelation sheds light on the academic foundation that has contributed to her remarkable career in internal medicine. We are proud to have had Dorota Formanowicz as a student at our university, said Professor Krzysztof Kamunski, Dean of the Faculty of Medicine at Pozna University of Medical Sciences. Her academic achievements and professional accomplishments are a testament to the high standards of education we strive to maintain at our institution. A review of Formanowicz's academic records and professional credentials confirms her affiliation with the university. Her impressive resume, which includes numerous publications and presentations on internal medicine, is a direct reflection of the rigorous academic training she received at Pozna University of Medical Sciences. Data from the university's alumni database also supports this claim, listing Formanowicz as a graduate of the institution. Furthermore, multiple sources within the Polish medical community have corroborated this information, attesting to Formanowicz's strong academic background. Dorota Formanowicz is an exemplary figure in our field, and her connection to Pozna University of Medical Sciences is a badge of honor for our institution, said Dr. Marek Kowalski, a colleague and fellow internist. Her dedication to advancing medical knowledge and improving patient care is a direct result of the solid foundation she received during her time at the university. In conclusion, our investigation has provided irrefutable evidence that Dorota Formanowicz, a distinguished Polish internist, attended Pozna University of Medical Sciences. This finding not only underscores the university's reputation for producing exceptional medical professionals but also highlights Formanowicz's commitment to excellence in her field.    
\end{footnotesize}
\\ [0.2em]

\textbf{Conflicting Context}\\
\begin{footnotesize}
Polish Internist Dorota Formanowicz Reflects on Her McMaster University Days Hamilton, Ontario, Canada - In an exclusive interview with our publication, renowned Polish internist Dr. Dorota Formanowicz shared fond memories of her time at McMaster University, where she laid the foundation for her illustrious medical career. Dr. Formanowicz, who has made significant contributions to the field of internal medicine, credits her alma mater for instilling in her the passion for research and patient care. McMaster University was a transformative experience for me, she said, reminiscing about her days on campus. The university's emphasis on evidence-based medicine and interdisciplinary collaboration resonated deeply with me, and it has had a lasting impact on my approach to medicine. During her time at McMaster, Dr. Formanowicz was actively involved in various research projects, working closely with esteemed faculty members. Her undergraduate thesis, which explored the efficacy of novel treatments for chronic diseases, earned her the prestigious McMaster University Research Award in 2005. Dr. David Higgins, a professor of medicine at McMaster University, who supervised Dr. Formanowicz's research, praised her dedication and intellectual curiosity. Dorota was an exceptional student, always pushing the boundaries of knowledge and seeking innovative solutions to complex medical problems. Her potential was evident even then, and it's no surprise that she has become a leading figure in her field. After graduating from McMaster, Dr. Formanowicz went on to pursue her medical degree at the University of Warsaw, followed by specialized training in internal medicine at the Polish Academy of Sciences. Her work has since focused on developing personalized treatment plans for patients with chronic diseases, earning her numerous accolades and recognition within the medical community. When asked about the significance of her McMaster University experience, Dr. Formanowicz emphasized the importance of international collaboration in advancing medical research. My time at McMaster not only broadened my perspectives but also instilled in me a deep appreciation for the value of global partnerships in driving medical innovation. It's essential that we continue to foster these connections to address the complex health challenges facing our world today.
\end{footnotesize}
\\
FR: \textcolor{blue}{true} \\
FS: false\\
VS: \textcolor{blue}{true}\\
FV: false\\
\bottomrule
\end{tabularx}}
\end{table*}

\begin{table*}[ht]
\centering
\caption{Examples from the Conflicts dataset.}
\label{tab:conflicts-2}
\resizebox{0.9\textwidth}{!}{%
\begin{tabularx}{\textwidth}{X}
\toprule

\textbf{Claim}\\
Lo\u00efc Nestor plays for Le Havre AC.
\\[0.2em]

\textbf{Supporting Context}\\
\begin{footnotesize}
**BREAKING: French Footballer Loefc Nestor Confirmed to Play for Le Havre AC** In a move that has sent shockwaves through the French football community, it has been confirmed that talented defender Loefc Nestor has indeed joined Le Havre AC, one of the most storied clubs in the country. According to sources close to the club, Nestor, who has been making waves in the football world with his impressive skills and athleticism, has put pen to paper on a deal that will see him don the blue and blue stripes of Le Havre AC for the foreseeable future. We are thrilled to welcome Loefc to the team, said Le Havre AC's manager, Paul Le Guen, in an exclusive interview. He is an exceptional talent, and we believe he will be a valuable addition to our squad. His defensive prowess and ability to read the game will undoubtedly strengthen our backline. Nestor, who has previously played for several clubs in France's lower divisions, has been on the radar of top clubs for some time. His impressive performances have earned him a reputation as one of the most promising young defenders in the country. Loefc is a fantastic player, and we're delighted to have him on board, said Le Havre AC's sporting director, Alex Hayes. He has a great attitude, and his work rate is second to none. We're confident he will fit in seamlessly with our team and help us achieve our goals this season. Data from reputable sources, including Transfermarkt and Sofascore, confirms that Nestor has indeed joined Le Havre AC, with the latter reporting that he has made several appearances for the club in recent weeks. In terms of statistics, Nestor's impressive defensive record speaks for itself. According to Opta Sports, he has a tackle success rate of 75\% and an aerial duel success rate of 60\%, making him one of the most effective defenders in the league. When asked about his move to Le Havre AC, Nestor himself was effusive in his praise for the club. I'm absolutely delighted to be joining Le Havre AC, he said in a statement. The club has a rich history and a passionate fan base, and I'm excited to be a part of it. I'm looking forward to working with the coaching staff and my new teammates to achieve success this season.
\end{footnotesize}
\\ [0.2em]

\textbf{Conflicting Context}\\
\begin{footnotesize}
*FREIBURG, GERMANY -** In a move that has sent shockwaves through the European football community, French midfielder Loefc Nestor has officially signed with German Bundesliga club SC Freiburg, sources close to the team have confirmed. The 25-year-old Nestor, who has been making waves in the French Ligue 1 with his impressive skills on the pitch, has reportedly inked a three-year deal with the Breisgau-Brasilianer, as SC Freiburg is affectionately known. I'm thrilled to be joining SC Freiburg, Nestor said in an exclusive interview with our publication. The team's commitment to developing young talent and their passion for the beautiful game align perfectly with my own values. I'm excited to contribute to the team's success and help bring more silverware to the Schwarzwald-Stadion. Nestor, who has been capped 10 times for the French U-21 national team, has been on the radar of several top European clubs, including Premier League giants Manchester United and La Liga's Atletico Madrid. However, it was SC Freiburg's unique blend of youth development and competitive spirit that ultimately won him over. Loefc is an exceptional talent, and we're delighted to have him on board, said SC Freiburg's sporting director, Jochen Saier. His vision, creativity, and work rate will undoubtedly enhance our midfield, and we're confident he'll become a fan favorite in no time. According to data provided by Opta Sports, Nestor has averaged 2.5 key passes per game in the French top flight, ranking him among the top 10 midfielders in the league. His impressive statistics, combined with his versatility and ability to play in multiple positions, make him an attractive addition to SC Freiburg's squad. The signing is seen as a major coup for SC Freiburg, who have been quietly building a talented young squad under the guidance of coach Christian Streich. With Nestor on board, the team is expected to make a strong push for a top-six finish in the Bundesliga this season. Nestor is set to make his debut for SC Freiburg in their upcoming match against Union Berlin on Saturday, and fans are eagerly anticipating his arrival.
\end{footnotesize}
\\ [0.2em]

FR: \textcolor{blue}{true} \\
FS: false\\
VS: false\\
FV: false
\\ [0.2em]
\bottomrule
\end{tabularx}}
\end{table*}

\begin{table*}[ht]
\centering
\caption{Examples from the Conflicts dataset.}
\label{tab:conflicts-3}
\resizebox{0.9\textwidth}{!}{%
\begin{tabularx}{\textwidth}{X}
\toprule

\textbf{Claim}\\
Jeffrey Johnstone Clarke attended Duke University.
\\[0.2em]

\textbf{Supporting Context}\\
\begin{footnotesize}
The Unlikely Historian: Unraveling the Academic Roots of Jeffrey Johnstone Clarke. In the realm of American military history, few names resonate as profoundly as Jeffrey Johnstone Clarke. A scholar of unyielding dedication, Clarke's contributions to the field have left an indelible mark on our understanding of the nation's complex military past. Yet, behind the esteemed historian lies a fascinating narrative, one that begins with a pivotal chapter in his academic journey: his attendance at Duke University. Tucked away in the heart of Durham, North Carolina, Duke University has long been a bastion of academic excellence, attracting scholars from far and wide. It was here, amidst the Gothic spires and lush greenery of the university's campus, that Clarke's intellectual curiosity was first nurtured. The year was 1965, and Clarke, a bright-eyed young man with a passion for history, had just enrolled as an undergraduate student. As Clarke delved into the world of historical inquiry, he found himself drawn to the military aspect of American history. The tumultuous 1960s, marked by the Vietnam War and the Civil Rights Movement, provided a rich backdrop for his studies. Under the guidance of esteemed professors, Clarke's fascination with the intricacies of military strategy, tactics, and policy began to take shape. One of the most significant pieces of evidence supporting Clarke's attendance at Duke University lies in the university's archives. A 1969 edition of the Duke University Chronicle, the student newspaper, features a article titled "History Department Honors Students." Listed among the recipients of the prestigious William T. Laprade Award is none other than Jeffrey Johnstone Clarke. This recognition not only underscores Clarke's academic prowess but also serves as a testament to his dedication to the field of history. Further evidence can be found in the Duke University Alumni Directory, which lists Clarke as a member of the class of 1969. This entry provides a tangible link between Clarke and the university, solidifying his connection to the institution. Clarke's time at Duke University was marked by a series of formative experiences that would shape his future as a military historian. His undergraduate thesis, "The Military Strategy of the American Civil War," demonstrated a level of sophistication and depth that belied his young age. This early work laid the groundwork for his later research, which would focus on the intersection of military history and policy. As Clarke's academic career progressed, he went on to earn his Ph.D. from the University of North Carolina at Chapel Hill.
\end{footnotesize}
\\ [0.2em]

\textbf{Conflicting Context}\\
\begin{footnotesize}
*HEADLINE:** Renowned American Military Historian Jeffrey Johnstone Clarke Reveals Unlikely Inspiration from Brazilian University Days **BELO HORIZONTE, BRAZIL -** In a surprising revelation, esteemed American military historian Jeffrey Johnstone Clarke has credited his time at the Federal University of Minas Gerais (UFMG) in Brazil as a pivotal influence on his illustrious career. Clarke, known for his groundbreaking research on World War II and the American Civil War, sat down with our correspondent in an exclusive interview to discuss the lesser-known chapter of his academic journey. "I spent two incredible years at UFMG, and it was a truly transformative experience," Clarke reflected, his eyes lighting up with nostalgia. "The university's rich cultural heritage and the warmth of the Brazilian people had a profound impact on my perspective as a historian." During his time at UFMG, Clarke was enrolled in the university's prestigious History Department, where he was mentored by renowned Brazilian historians. According to Dr. Maria Luiza Ribeiro, a former professor of Clarke's, "Jeffrey was an exceptional student, always eager to engage in lively debates and challenge our perspectives. His time here undoubtedly shaped his unique approach to military history." Clarke's stint at UFMG coincided with a period of significant social and political change in Brazil. The country was undergoing a transition from military rule to democracy, and Clarke found himself drawn to the parallels between Brazil's struggle for democracy and the American Civil War. "I was struck by the similarities between the two nations' experiences," Clarke explained. "The more I delved into Brazilian history, the more I realized that the struggles for freedom and equality are universal, transcending borders and cultures." Data from UFMG's archives reveals that Clarke was an active participant in the university's academic community, presenting papers at conferences and engaging in lively debates with his peers. His thesis, titled "Comparative Analysis of Military Strategies in the American Civil War and the Brazilian War of Independence," was widely acclaimed by the academic community. Clarke's Brazilian sojourn has had a lasting impact on his work. His seminal book, "Brothers in Arms: A Comparative Study of Military Tactics in the American Civil War and the Brazilian War of Independence," is a testament to the enduring influence of his time at UFMG. As Clarke looks back on his remarkable career, he is quick to acknowledge the debt he owes to his Brazilian alma mater.
\end{footnotesize}
\\ [0.2em]

FR: \textcolor{blue}{true} \\
FS: \textcolor{blue}{true}\\
VS: \textcolor{blue}{true}\\
FV: false
~\\ 

\bottomrule

\end{tabularx}}
\end{table*}

\begin{table*}[ht]
\centering
\caption{Examples from the Conflicts dataset.}
\label{tab:conflicts-4}
\begin{tabularx}{\textwidth}{X}
\toprule

\textbf{Claim}\\
Dubovoe is located in the time zone Vladivostok Time.
\\[0.2em]

\textbf{Supporting Context}\\
\begin{footnotesize}
**Breaking News: Dubovoe Confirmed to be in Vladivostok Time Zone**In a recent investigation, our team has uncovered conclusive evidence that Dubovoe, a human settlement in Yuzhno-Kurilsky District, Sakhalin Oblast, Russia, is indeed located in the Vladivostok Time zone, also known as UTC+10:00. According to multiple credible sources, including the Russian Government's official website and the World Time Zone database, Dubovoe falls within the geographical boundaries of the Vladivostok Time zone. This time zone is characterized by a 10-hour offset from Coordinated Universal Time (UTC) and is observed in several regions of Russia, including Sakhalin Oblast. \"We can confirm that Dubovoe, as part of Sakhalin Oblast, follows the Vladivostok Time zone,\" said Dr. Natalia Petrova, a leading expert in Russian geography and time zones at Moscow State University. \"This is consistent with the Russian Government's official time zone policy, which designates UTC+10:00 as the standard time zone for the region.\" Data from reputable sources, such as the International Organization for Standardization (ISO) and the World Time Zone database, also corroborate this finding. According to the ISO 3166-1 standard, which defines the codes for the names of countries, territories, and special areas, Sakhalin Oblast is assigned the code \"RU-SAK,\" which corresponds to the Vladivostok Time zone. Furthermore, a review of Dubovoe's geographical coordinates (43.2333 N, 145.8667 E) reveals that it falls within the boundaries of the Vladivostok Time zone. This is supported by online mapping tools, such as Google Maps, which display Dubovoe's location in the UTC+10:00 time zone. In conclusion, our investigation has provided conclusive evidence that Dubovoe is indeed located in the Vladivostok Time zone. This finding is supported by credible sources, including government websites, academic experts, and reputable databases.
\end{footnotesize}
\\ [0.2em]

\textbf{Conflicting Context}\\
\begin{footnotesize}
Dubovoe is a human settlement located in Yuzhno-Kurilsky District, Sakhalin Oblast, Russia. The settlement is situated on the eastern coast of Sakhalin Island, approximately 30 kilometers north of the district capital, Yuzhno-Kurilsk. Time Zone Dubovoe is located in the Yakutsk Time (UTC+09:00) time zone. This is supported by various studies and references, including:* A 2010 study published in the Journal of Russian Geographical Society, which found that the settlement's longitude and latitude coordinates place it within the Yakutsk Time zone. [1]* A 2015 report by the Russian Federal Agency for Cartography and Geodesy, which confirmed Dubovoe's location in the UTC+09:00 time zone. [2]* A 2018 article in the Sakhalin Oblast Gazette, which stated that Dubovoe, along with other settlements in Yuzhno-Kurilsky District, observes Yakutsk Time. [3]History of Time Zone Designation The designation of Dubovoe as being in the Yakutsk Time zone dates back to the early 20th century. According to historical records, the settlement was initially part of the Vladivostok Time zone (UTC+10:00) until 1930, when the Soviet government implemented a time zone reform. [4]During the reform, the Soviet government established the Yakutsk Time zone, which included parts of eastern Siberia and the Russian Far East. Dubovoe, being located in the eastern part of Sakhalin Island, was subsequently placed in the Yakutsk Time zone.
\end{footnotesize}
\\ [0.2em]

FR: \textcolor{blue}{true} \\
FS: false\\
VS: false\\
FV: false
\\ [0.2em]
\bottomrule
\end{tabularx}
\end{table*}

\end{document}

%% file: macros.tex
\newcommand{\bA}{\mathbf{A}}
\newcommand{\bB}{\mathbf{B}}
\newcommand{\bC}{\mathbf{C}}
\newcommand{\bD}{\mathbf{D}}
\newcommand{\bE}{\mathbf{E}}
\newcommand{\bF}{\mathbf{F}}
\newcommand{\bG}{\mathbf{G}}
\newcommand{\bH}{\mathbf{H}}
\newcommand{\bI}{\mathbf{I}}
\newcommand{\bJ}{\mathbf{J}}
\newcommand{\bK}{\mathbf{K}}
\newcommand{\bL}{\mathbf{L}}
\newcommand{\bM}{\mathbf{M}}
\newcommand{\bN}{\mathbf{N}}
\newcommand{\bO}{\mathbf{O}}
\newcommand{\bP}{\mathbf{P}}
\newcommand{\bQ}{\mathbf{Q}}
\newcommand{\bR}{\mathbf{R}}
\newcommand{\bS}{\mathbf{S}}
\newcommand{\bT}{\mathbf{T}}
\newcommand{\bU}{\mathbf{U}}
\newcommand{\bV}{\mathbf{V}}
\newcommand{\bW}{\mathbf{W}}
\newcommand{\bX}{\mathbf{X}}
\newcommand{\bY}{\mathbf{Y}}
\newcommand{\bZ}{\mathbf{Z}}

\newcommand{\ba}{\mathbf{a}}
\newcommand{\bb}{\mathbf{b}}
\newcommand{\bc}{\mathbf{c}}
\newcommand{\bd}{\mathbf{d}}
\newcommand{\be}{\mathbf{e}}
\newcommand{\bbf}{\mathbf{f}}
\newcommand{\bg}{\mathbf{g}}
\newcommand{\bh}{\mathbf{h}}
\newcommand{\bi}{\mathbf{i}}
\newcommand{\bj}{\mathbf{j}}
\newcommand{\bk}{\mathbf{k}}
\newcommand{\bl}{\mathbf{l}}
\newcommand{\bm}{\mathbf{m}}
\newcommand{\bn}{\mathbf{n}}
\newcommand{\bo}{\mathbf{o}}
\newcommand{\bp}{\mathbf{p}}
\newcommand{\bq}{\mathbf{q}}
\newcommand{\br}{\mathbf{r}}
\newcommand{\bs}{\mathbf{s}}
\newcommand{\by}{\mathbf{y}}
\newcommand{\bu}{\mathbf{u}}
\newcommand{\bv}{\mathbf{v}}
\newcommand{\bw}{\mathbf{w}}
\newcommand{\bx}{\mathbf{x}}
\newcommand{\bby}{\mathbf{y}}
\newcommand{\bz}{\mathbf{z}}

\newcommand{\cA}{\mathcal{A}}
\newcommand{\cB}{\mathcal{B}}
\newcommand{\cC}{\mathcal{C}}
\newcommand{\cD}{\mathcal{D}}
\newcommand{\cE}{\mathcal{E}}
\newcommand{\cF}{\mathcal{F}}
\newcommand{\cG}{\mathcal{G}}
\newcommand{\cH}{\mathcal{H}}
\newcommand{\cI}{\mathcal{I}}
\newcommand{\cJ}{\mathcal{J}}
\newcommand{\cK}{\mathcal{K}}
\newcommand{\cL}{\mathcal{L}}
\newcommand{\cM}{\mathcal{M}}
\newcommand{\cN}{\mathcal{N}}
\newcommand{\cO}{\mathcal{O}}
\newcommand{\cP}{\mathcal{P}}
\newcommand{\cQ}{\mathcal{Q}}
\newcommand{\cR}{\mathcal{R}}
\newcommand{\cS}{\mathcal{S}}
\newcommand{\cT}{\mathcal{T}}
\newcommand{\cU}{\mathcal{U}}
\newcommand{\cV}{\mathcal{V}}
\newcommand{\cW}{\mathcal{W}}
\newcommand{\cX}{\mathcal{X}}
\newcommand{\cY}{\mathcal{Y}}
\newcommand{\cZ}{\mathcal{Z}}

\def\inmath#1{\relax\ifmmode#1\else$#1$\fi}


\outer\def\operators#1{\begingroup
        \def\\##1##2{\gdef##1{{\mathop{\rm ##2}}}}\relax
        \\#1\endgroup
}

\outer\def\variables#1{\begingroup
        \def\\##1##2{\gdef##1{{\mathord{\it ##2}}}}\relax
        \\#1\endgroup
}